%% file: manuscript.tex
\documentclass{article}

\PassOptionsToPackage{numbers, compress}{natbib}

    \usepackage[preprint]{neurips_2023}

\usepackage[utf8]{inputenc} 
\usepackage[T1]{fontenc}    
\usepackage{hyperref}       
\usepackage{url}            
\usepackage{booktabs}       
\usepackage{amsfonts}       
\usepackage{nicefrac}       
\usepackage{microtype}      
\usepackage{xcolor}         

\usepackage{graphicx} 
\usepackage{subcaption}
\usepackage{multirow} 
\usepackage{rotating} 
\usepackage{amsmath}
\usepackage{amssymb}

\title{On the calibration of compartmental \\epidemiological models\thanks{\textit{Taken from N.G.'s thesis submitted to the Faculty of the New York University Tandon School of Engineering in partial fulfillment of the requirements for the degree Master of Science in Computer Engineering, New York University Tandon School of Engineering.}}}

\usepackage[textsize=footnotesize]{todonotes}

\author{%
Nikunj Gupta$^{1}$\thanks{All correspondence can be directed to \texttt{nikunj.gupta@nyu.edu}.} \quad Anh Mai$^{2}$ \quad Azza Abouzied$^{2}$ \quad Dennis Shasha$^{1}$\\
$^1$New York University \quad $^2$NYU Abu Dhabi \\
\texttt{\{nikunj.gupta,anh.mai,azza\}@nyu.edu, shasha@cims.nyu.edu}
}

\begin{document}

\maketitle

\begin{abstract}
  \input{abstract}
\end{abstract}


\section{Introduction}
\label{sec:introduction}
\input{chapters/01_introduction}


\section{Optimization methods for Model Calibration}
\label{sec:optim_methods}
\input{chapters/02_optim_methods}


\section{Dealing with noise in data}
\label{sec:results}
\input{chapters/03_noise_in_data} 

\section{Working with Population Subgroups}
\label{sec:population_subgroups}
\input{chapters/04_population_subgroups}


\section{Practical Conclusions}
\label{sec:conclusions}
\input{chapters/05_practical_conclusions}


\section{Reinforcement Learning}
\label{sec:rl}
\input{chapters/06_reinforcement_learning}


\section{Conclusion} 
\label{sec:conclusion} 
\input{chapters/08_Conclusion}

\begin{ack}
    \input{acknowledge} 
\end{ack}

{
\small
\bibliographystyle{plain} 
\bibliography{manuscript.bib} 
}

\appendix

\section{Appendix}

\input{chapters/10_appendix}

\end{document}

%% file: abstract.tex
Epidemiological compartmental models are useful for understanding infectious disease propagation and directing public health policy decisions. Calibration of these models is an important step in offering accurate forecasts of disease dynamics and the effectiveness of interventions. In this study, we present an overview of calibrating strategies that can be employed, including several optimization methods and reinforcement learning (RL). We discuss the benefits and drawbacks of these methods and highlight relevant practical conclusions from our experiments. Optimization methods iteratively adjust the parameters of the model until the model output matches the available data, whereas RL uses trial and error to learn the optimal set of parameters by maximizing a reward signal. Finally, we  discuss how the calibration of parameters of epidemiological compartmental models is an emerging field that has the potential to improve the accuracy of disease modeling and public health decision-making. Further research is needed to validate the effectiveness and scalability of these approaches in different epidemiological contexts. All codes and resources are available on \url{https://github.com/Nikunj-Gupta/On-the-Calibration-of-Compartmental-Epidemiological-Models}. We hope this work can facilitate related research. 

%% file: chapters/01_introduction.tex
Epidemiologists utilize mathematical models to better understand and forecast disease spread in communities, allowing them to replicate the complicated interactions between people, viruses, and environmental factors that drive disease transmission. They can then forecast the path of an outbreak, evaluate the performance of various interventions, and select the most effective disease-control techniques. Commonly used models include compartmental models, agent-based models, and network models. We use compartmental models, which are described in the next subsection. 

\begin{figure}[ht]
    \centering
    \includegraphics[scale=0.4, bb= 0 0 401 361]{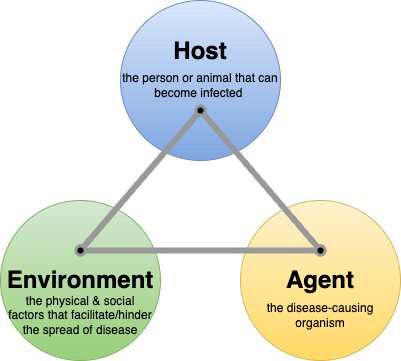}
    \caption{Epidemiologists frequently employ the shown \textbf{epidemiological triangle} framework of three interconnected components to discover the elements that contribute to disease transmission and to develop strategies to control or eliminate its spread.} 
    \label{fig:epidemiological_triangle}
\end{figure}

\subsection{Epidemic compartmental models}

\begin{figure}[ht]
    \centering
    \includegraphics[width=0.5\textwidth, bb= 0 0 441 61]{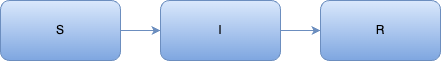}
    \caption{SIR model} 
    \label{fig:sir}
\end{figure}

Epidemic compartmental models are mathematical models that are used to examine how infectious illnesses propagate within a population. These models divide the population into compartments or groups depending on disease status and follow hosts' mobilities between these compartments over time. The SIR model, the SIRD model, and the SIRVD model are the most frequent forms of epidemic compartmental models (and the ones we primarily used in this study): 
 
\paragraph{SIR Model: \cite{kermark1927contributions,harko2014exact,beckley2013modeling,kroger2020analytical,schlickeiser2021analytical}} Susceptible (S), Infected (I), and Recovered (R) are the three compartments in the SIR model, presuming that individuals in the community are either susceptible to the disease, infected with the disease, or have recovered from the illness and established immunity. The model follows the movement of individuals from the susceptible to the infected compartment and from the infected to the recovered compartment based on parameters such as the transmission rate, recovery rate, and the number of susceptible persons in the population. Figure~\ref{fig:sir} shows a diagram for SIR model. 

\begin{figure}[ht]
    \centering
    \includegraphics[width=0.5\textwidth, bb= 0 0 441 181]{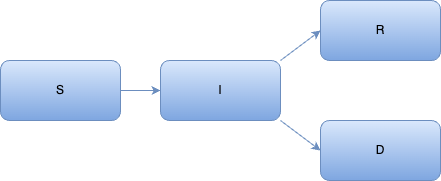}
    \caption{SIRD model} 
    \label{fig:sird}
\end{figure}

\paragraph{SIRD Model: \cite{bailey1975mathematical}} The SIRD model extends the SIR model by including a Deaths (D) compartment. This model follows the flow of individuals from the infected compartment to either the recovered or the dying compartment and can provide insights into the disease's mortality rate and overall influence on the population. Figure~\ref{fig:sird} shows a diagram for SIRD model.

\begin{figure}[ht]
    \centering
    \includegraphics[width=0.5\textwidth, bb= 0 0 441 241]{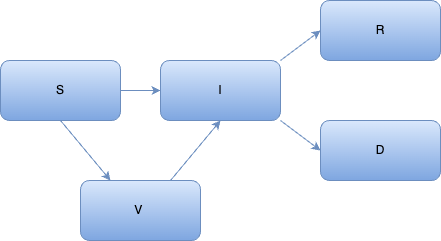}
    \caption{SIRVD model} 
    \label{fig:sirvd}
\end{figure}

\paragraph{SIRVD Model: \cite{farooq2020novel}} The SIRVD model extends on the SIRD model by including a compartment for Vaccinated (V). It enables the evaluation of the impact of vaccination on disease spread and can assist influence public health initiatives connected to vaccination campaigns, such as predicting the coverage required to obtain herd immunity or evaluating the success of various vaccination strategies. Figure \ref{fig:sirvd} depicts a diagram for the SIRVD model. 

These compartmental models have been widely utilized to simulate infectious disease propagation, quantify the impact of interventions, and guide public health policies. They are based on over-simplified but still useful assumptions about disease dynamics and population characteristics. 

\subsection{Calibration of epidemic compartmental models} 

The compartmental models depend on a number of parameters that represent illness patterns, population characteristics, and treatments such as vaccination or social distancing measures. Such model parameters need to be fitted (or calibrated) to real-world data to improve the accuracy and relevance of model predictions. Then, such a calibrated model can be used to simulate and predict disease spread under various scenarios. However, it can be challenging to accurately calibrate epidemic compartmental models, due to uncertainties in data quality, model assumptions, and even parameter values. It is also critical to iterate and refine the model as new data becomes available or as the scenario evolves. 


\subsection{Problem Statement} 

Because of the recent global health crises due to COVID-19, being able to mimic and hence predict the evolution of an epidemic has become a major concern. The aim of this manuscript is to investigate several machine learning and optimization-based methods for estimating the parameters of  compartmental epidemiological models. We investigate and compare performances using a variety of \textit{ordinary differential equation} (ODE)-based simulated datasets and models such as SIR, SIRD, and SIRVD. The goal of this study is to be able to determine the best calibration strategy given a disease dynamic
model (e.g. SIRVD) and characteristics of the data (e.g. dataset size or noise characteristics). 


%% file: chapters/02_optim_methods.tex
The use of optimization methods to search for the parameter values that best fit the available data is a practical and common strategy for calibrating epidemic models. They can account for the complex interactions between multiple parameters while estimating them. Some common approaches include the following: 

\paragraph{Gradient-based methods:} These approaches compute the gradient of the cost function with respect to the model parameters and update the values in the direction of the steepest descent. The Levenberg-Marquardt algorithm is a well-known gradient-based method for nonlinear least-squares optimization. 

\paragraph{Gradient-free methods:} When the cost function is non-differentiable or the gradient of the cost function is difficult to compute, gradient-free methods such as the Nelder-Mead algorithm, Powell's approach, and the differential evolution algorithm can be effective. These methods improve the cost function by iteratively searching the parameter space and modifying their values. These approaches are generally less efficient than gradient-based methods, but they can be effective in locating global optima in large search areas. 

\paragraph{Simulated annealing:} This is a stochastic optimization method inspired by the metallurgical annealing process. It works by perturbing parameter values iteratively and accepting or rejecting new values depending on a probability distribution that changes over time as a metaphorical temperature value cools. At higher temperatures, value changes are likely to be taken even if they decrease some goodness score. That likelihood decreases as the temperature goes down. Such an approach can be beneficial for locating global optima in complicated and multi-modal search spaces, but it is slow to converge and may require a significant number of function evaluations. 

Before discussing the optimization methods in  detail, we first discuss the objective function and the (simulated) data used in this article. 

\subsection{The objective function}

In order to obtain an estimation of the parameters, we consider the model derived from the data to be a  compartmental model (i.e. SIR, SIRD, etc.). We will then try to find the set of parameters that will minimize the following function:

\begin{center}
   $\sum_{c} \sum_{days }(actual - expected)^2$
\end{center}

with c being the compartments available for a fit. By \textit{actual} we mean data resulting from the simulation and by \textit{expected} we mean measured data. We also note here that data for some  compartments may neither be available nor necessary in order to obtain a correct fit. 

\subsection{Description of simulated data} 
\label{sec:simulated_data}

In this study, we simulate data for all the aforementioned compartmental models (SIR, SIRD, SIRVD), by tracking the number of individuals in each compartment at each time step and by updating these values as time progresses. To simulate a compartmental model without code, we took the following steps: 

\begin{enumerate}
    
    \item Set initial conditions: the number of individuals in each compartment at the beginning of the simulation. For example, to simulate an outbreak of a new disease in a population of 10,000 people we start with 9,999 susceptible individuals and 1 infectious individual. 

    \item Determine model Parameters: the rates at which individuals move between compartments. For instance, in the SIR model, these parameters are the infection rate (beta) and the recovery rate (gamma). 

    \item Set time step: we chose one day as the time interval for our model simulations. 

    \item Track and calculate values for each compartment: For each time step, calculate the new values for each compartment using the equations of the model. For the SIR model, these equations are: 

    \[ \frac{\partial S}{\partial t} = -\beta SI \]
    \[ \frac{\partial I}{\partial t} = \beta SI - \gamma I \]
    \[ \frac{\partial R}{\partial t} = \gamma I \] 
    
    where, t is the timeline, and S, I, and R are the number of individuals in the respective compartments. $\frac{\partial S}{\partial t}$ denotes the rate of change of S with respect to t (similarly for all compartments). 
    
    \item Repeat step 5 for each time step, tracking/updating the values at each step. 
    
    \item Plot the results using a line graph, with time on the x-axis and the number of individuals in each compartment on the y-axis. 
    
\end{enumerate}

Simulating compartmental models like this can be a helpful exercise to gain a better understanding of the dynamics of infectious disease outbreaks and other complex systems. However, for more complex models or larger populations, we use   more efficient code and specialized software (Epipolicy) for simulation\footnote{Discussed in detail later in Section~\ref{sec:population_subgroups}}. 

In the next subsection, we  describe all the optimization methods we used for this study in more detail. 

\subsection{Description of methods}


\paragraph{The Levenberg-Marquardt algorithm \cite{more2006levenberg}.} This algorithm is used to solve nonlinear least squares problems. The idea is to iteratively update the model parameters and reduce the difference between the predicted data points (using estimated model parameters) and true values of the number of individuals in each compartment from a simulated/available epidemic dataset.
It starts with a plausible guess 
for the parameters and then uses gradient descent to update them. 

\paragraph{Least-Squares minimization using \textit{Trust Region Reflective} method \cite{branch1999subspace}.} It uses a trust region strategy over least-squares minimization to ensure that the updates stay within a predefined region of the parameter space. First, we compute the Jacobian matrix (the partial derivatives of the objective function
with respect to each parameter) and use it to estimate the curvature of the objective function as well as determine the size of the trust region. Next, the parameters are iteratively updated by minimizing the objective function within the trust region. If the objective function decreases, the trust region is increased, allowing the algorithm to explore a larger portion of the parameter space, and vice-versa if the function increases. This method is robust, however, it can be sensitive to the choice of initial parameters and can suffer from slow convergence for poorly conditioned problems. Additionally, computing the Jacobian matrix can be computationally expensive for high-dimensional problems. 

\paragraph{Differential Evolution (DE) \cite{storn1997differential}.} DE iteratively improves a population of candidate solutions to find the best solution for a problem. It uses random perturbations, called mutations, to generate trial solutions from the population. These trial solutions are then combined with the original population through crossover, resulting in a new population. The best solutions from the new population are then selected using an objective function, and the process repeats for a certain number of iterations or until a termination condition is met. DE is known for its simplicity and ability to find optimal solutions in various real-world problems. 

\paragraph{Brute force optimization.}\footnote{Implemented using Scipy (\url{https://docs.scipy.org/doc/scipy/reference/generated/scipy.optimize.brute.html})} Without employing any specific algorithm or heuristics, this method exhaustively evaluates all possible solutions within a specified search space to find the one that optimizes the objective function. It is straightforward to implement but is computationally expensive for large search spaces. However, it guarantees to find the global optimal solution, if it exists within some discrete search space. 

\paragraph{The basin-hopping algorithm \cite{olson2012basin}.} This is like searching for the lowest point in a landscape with valleys and hills. It starts from an initial point and makes small random moves to nearby points. Then it looks around those points to see if there's a lower point nearby. If it finds a lower point, it moves there and repeats the process. It keeps doing this, occasionally making bigger random moves to explore new areas, until it can't find any lower points or a stopping condition is met. The idea is to keep searching for the global minimum, hopping out of local valleys (or basins) to explore new areas and find the overall lowest point in the landscape. It is used for finding the best solution in complex optimization problems with many peaks and valleys. This can be viewed as a primitive version of simulated annealing.

\paragraph{Adaptive Memory Programming (AMP) for Global Optimization \cite{lasdon2010adaptive}.} Adaptive Memory Programming (AMP) is a metaheuristic algorithm for global optimization that uses an adaptive memory mechanism to guide the search process. It maintains a set of elite solutions, called the memory, which contains the best solutions found so far. The search process starts with an initial population of solutions, which is generated randomly or using other methods such as local search. The solutions are evaluated using an objective function, and the best ones are selected to form the initial memory. In each iteration, the algorithm generates a new population of solutions by combining the memory with the current population. The new solutions are then evaluated, and the best ones are added to the memory, replacing the worst ones if necessary. The adaptive memory mechanism in AMP plays a crucial role in the algorithm's ability to efficiently explore the search space, as it adjusts the balance between the exploitation of the best solutions in the memory and the exploration of the search space (by controlling the intensity of the search, the size of the memory, and the way the memory is updated). AMP's effectiveness comes from its ability to exploit the promising regions of the search space while maintaining diversity in the population, leading to high-quality solutions with a reasonable computational effort. 

\paragraph{Nelder-Mead optimization \cite{gao2012implementing}.} Also known as the Simplex method, the Nelder-Mead approach works by iteratively updating a simplex, which is a set of points in N-dimensional space until a satisfactory solution is found. It is a simple yet effective algorithm and is useful for optimizing functions in multi-dimensional space. It does not require the gradient or Hessian information of the objective function, making it suitable for a wide range of optimization problems. 

\paragraph{Powell's method \cite{powell1964efficient}.} This approach is a derivative-free, direct search method that also does not require the gradient or Hessian information of the objective function. It iteratively updates the search directions based on past iterations, using a combination of line searches and pattern moves. More specifically, it works by exploring the search space along different orthogonal directions, called \textit{Powell} directions to efficiently converge to the optimal solution. The method is known for its simplicity and robustness, however, it may not always converge as fast as gradient-based methods, especially for high-dimensional problems. 

\paragraph{Conjugate Gradient (CG) \cite{hestenes1952methods}.} CG is an iterative method that efficiently utilizes the gradient information of the objective function to search for the optimal solution along \textit{conjugate} directions (which ensures that the search directions are orthogonal to each other). It is a popular choice for optimizing large-scale problems due to its minimal memory and computation requirements. However, it requires a differentiable objective function which might not always be available. 

\paragraph{BFGS \cite{byrd1995limited}.} Broyden-Fletcher-Goldfarb-Shanno (BFGS) belongs to the class of quasi-Newton methods, which are iterative methods that approximate the Hessian matrix of the objective function to efficiently find the optimal solution without computing the actual Hessian. BFGS updates the approximation of the Hessian matrix based on the gradient information and the change in gradient between iterations and uses this updated approximation to determine the search direction and step size for optimization. It is known for its efficiency in finding the optimal solution for smooth, well-behaved objective functions, but, it may not be suitable for optimizing non-differentiable or poorly behaved objective functions. 

\paragraph{Limited-memory BFGS \cite{zhu1997algorithm}.} L-BFGS extends BFGS where the "limited-memory" part refers to the fact that it  stores only a limited amount of past information to approximate the Hessian matrix, making it memory-efficient for large-scale optimization problems. L-BFGS is known for its fast convergence. 

\paragraph{Trust-region for constrained optimization \cite{conn2000trust}.} This optimization uses a trust-region approach to handle constraints while searching for the optimal solution. It iteratively updates the search direction ensuring that these steps are taken within a certain region satisfying the constraints. This is typically useful for problems with both equality and inequality constraints and is known for its ability to handle constrained optimization problems efficiently and accurately. 

\paragraph{Truncated Newton (TN) \cite{nash2000survey}} TN is a variant of Newton's method that  computes only a partial or truncated version of the Hessian matrix to reduce the computational cost and uses it to update the search direction. It also uses a line search procedure to determine the step size along the search direction. TN is known for its efficiency in handling large-scale optimization problems, especially those with sparse Hessian matrices, and it can also handle non-linear constraints. However, TN may require more iterations to converge compared to other optimization algorithms and may be sensitive to the choice of the truncation parameter. 

\paragraph{Sequential Linear Squares Programming (SLSQP) \cite{kraft1988software}.} SLSQP is a sequential approach that iteratively solves a series of linearized subproblems to approximate the original nonlinear problem. These linear programming problems are then solved using suitable solvers, such as the Simplex method or an interior-point method. SLSQP is known for its effectiveness in handling nonlinear optimization problems with constraints. 

\paragraph{Simplicial homology global optimization (SHGO) \cite{endres2018simplicial}.} SHGO involves constructing a simplicial complex from the domain, which is a combinatorial structure composed of simplices of varying dimensions. By analyzing the topological properties of this complex, such as connected components, holes, and voids, it identifies regions of interest in the domain and guides the search for the global optimum. SHGO is known for its ability to handle complex and multimodal optimization problems. 

\paragraph{Dual Annealing optimization \cite{xiang1997generalized}.} This method combines the concepts of simulated annealing and genetic algorithms. It employs a dual approach where multiple optimizations occur in parallel, one at a local level and the other at a global level. Local optimization is performed using traditional optimization techniques, while global optimization involves random perturbations and a selection of candidate solutions based on a probability distribution. This approach is particularly effective for optimizing functions with multiple local optima, as it balances local and global exploration to search for the global optimum in a robust and efficient manner. 

From among the aforementioned methods\footnote {\textit{Lmfit}\footnote{\url{https://lmfit.github.io/lmfit-py/fitting.html}} is a Python library that provides a high-level interface for performing nonlinear optimization and curve fitting, making it a popular choice for calibrating compartmental epidemiological models. It includes all the aforementioned optimization algorithms (and more). The library is open-source and available on Github, making it easily accessible to researchers and practitioners in the field of epidemiology.}, Least-squares (Levenberg-Marquardt algorithm), Nelder-Mead, Powell's method, and Trust-region-based constrained optimization prove to be the most successful in many of our experiments (discussed more in detail in the following chapters). Also, in future work, it could be interesting to combine differential evolution (also known as genetic algorithms) with simulated annealing and either gradient evolution or simplex. 

\subsection{Discussion of calibration experiments and empirical results} 

In this subsection, we discuss and compare the performance of all the described methods on different models and datasets. The metric to compare the effectiveness of these methods is the Mean Absolute Error (MAE) between the predicted and true values of the number of people in the \textit{Infected} compartment. So, here, \textit{the lower the MAE, the better the curve fits the data}. 

The workflow of our experiments done as part of this study is as follows:

\begin{enumerate}

    \item Generate epidemic data using the steps mentioned in subsection \ref{sec:simulated_data}. 

    \item Pick out the first few days' data points to be given as input to the optimization method. 

    \item Fit model parameters using each optimization method on the picked data points. 

    \item Use the estimated model parameters to generate a curve for the entire timeline of the epidemic. 

    \item Calculate the MAE between the predicted curve and the true values corresponding to the number of people in the \textit{Infected} compartment. 

    \item Choose the best (or top-3) performing optimization methods for the given scenario. 

\end{enumerate}

\begin{figure}
\centering
    \begin{subfigure}[b]{\textwidth}
        \centering
        \includegraphics[width=0.3\linewidth, bb= 0 0 826 826]{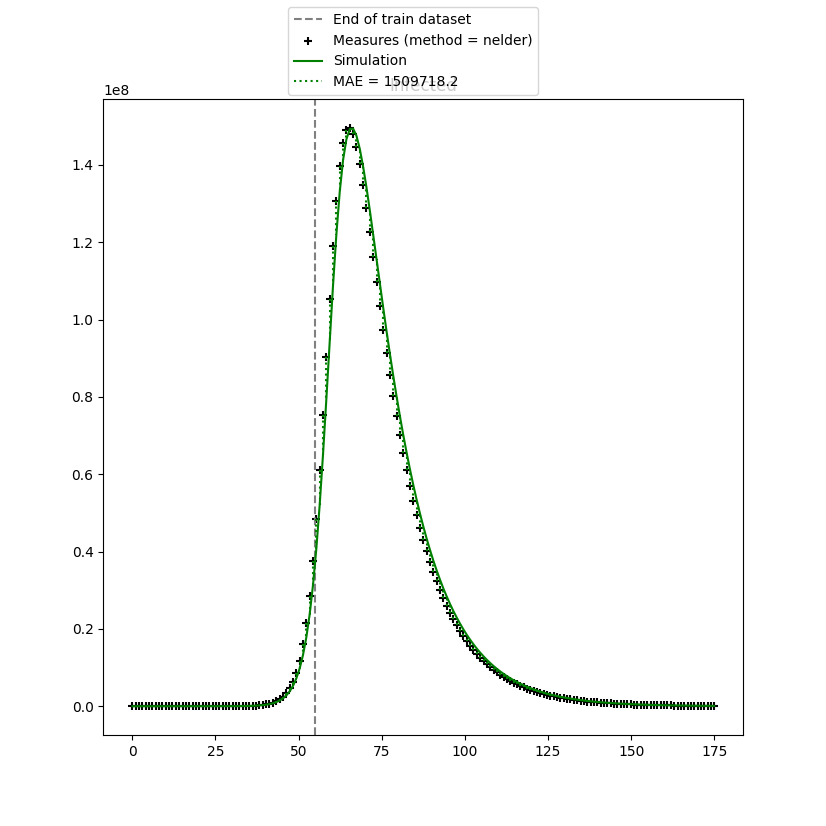}
        \includegraphics[width=0.3\linewidth, bb= 0 0 826 826]{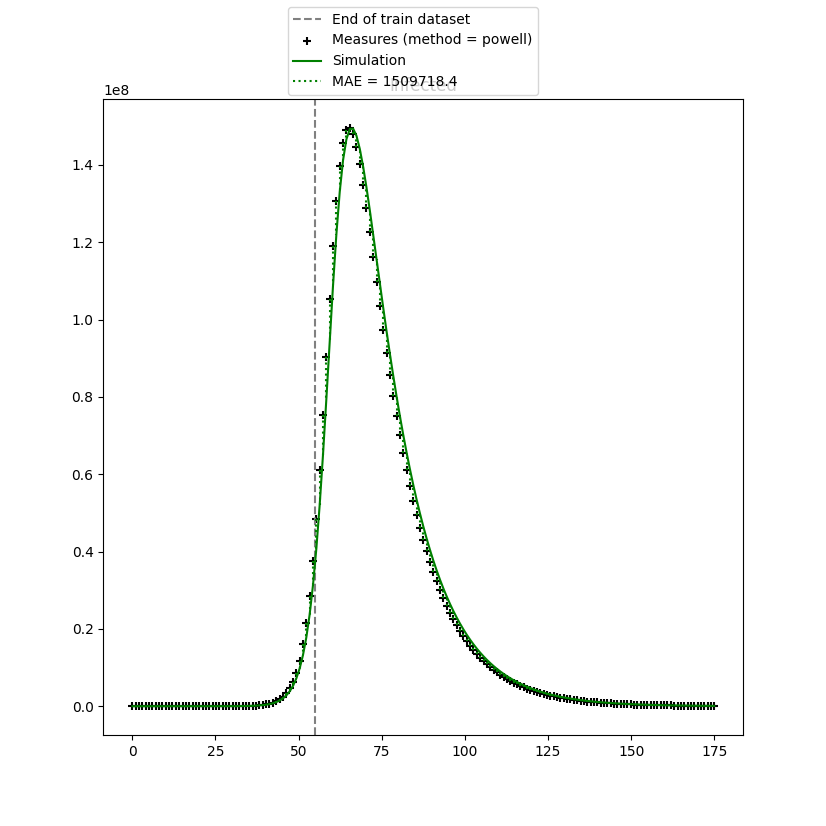}
        \includegraphics[width=0.3\linewidth, bb= 0 0 826 826]{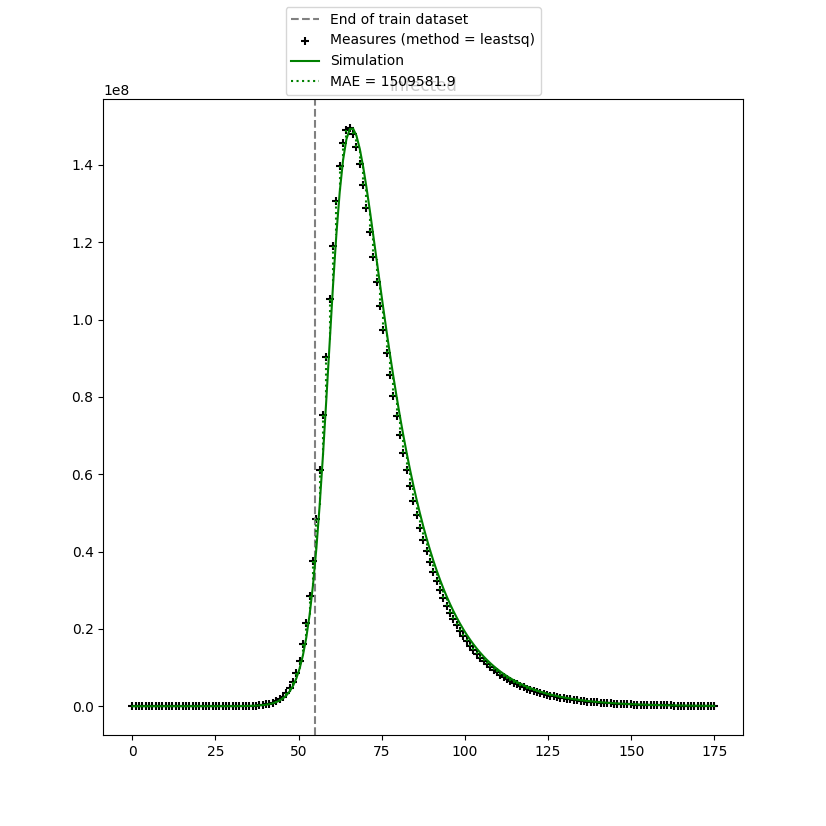}
        \caption{ SIR }
        \label{fig:top3-sir-lowdata-nonoise-nosubgroups}
    \end{subfigure} 

    \begin{subfigure}[b]{\textwidth}
        \centering
        \includegraphics[width=0.3\linewidth, bb= 0 0 826 826]{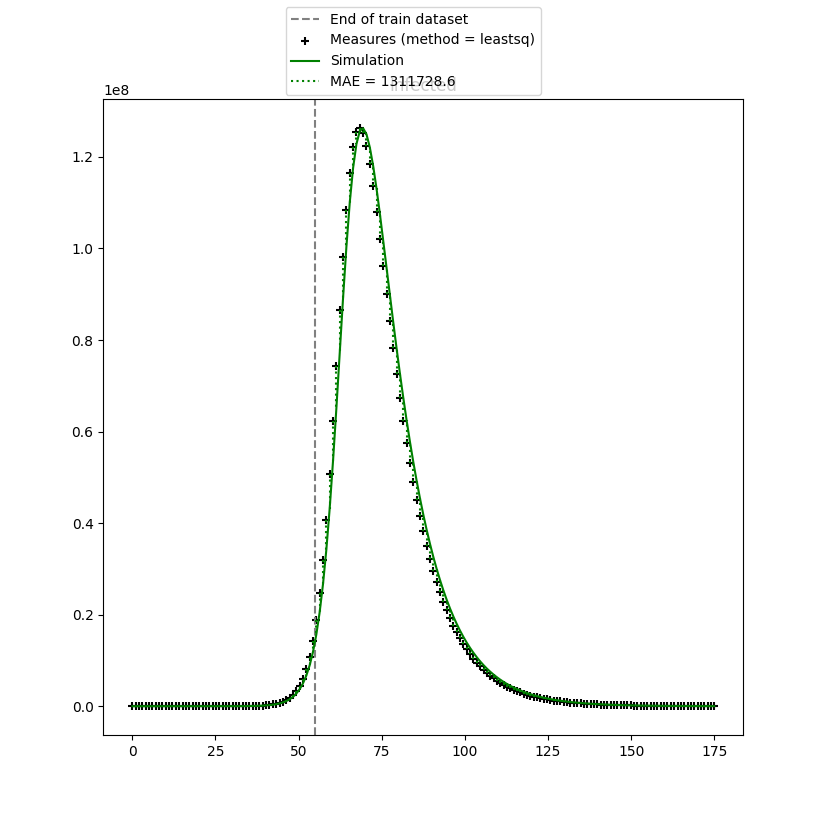}
        \includegraphics[width=0.3\linewidth, bb= 0 0 826 826]{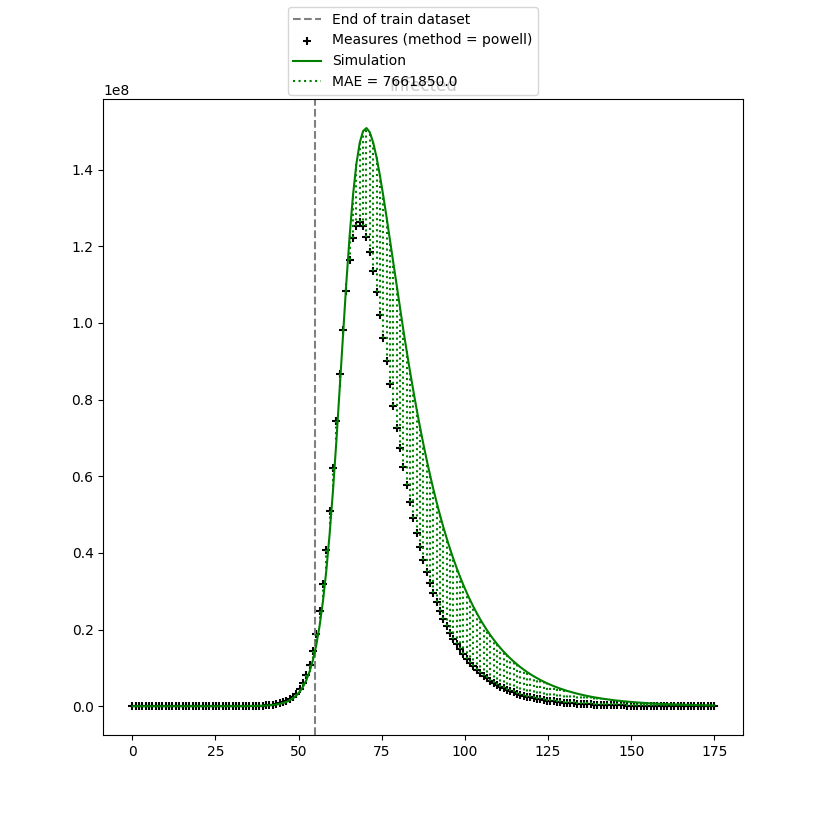}
        \includegraphics[width=0.3\linewidth, bb= 0 0 826 826]{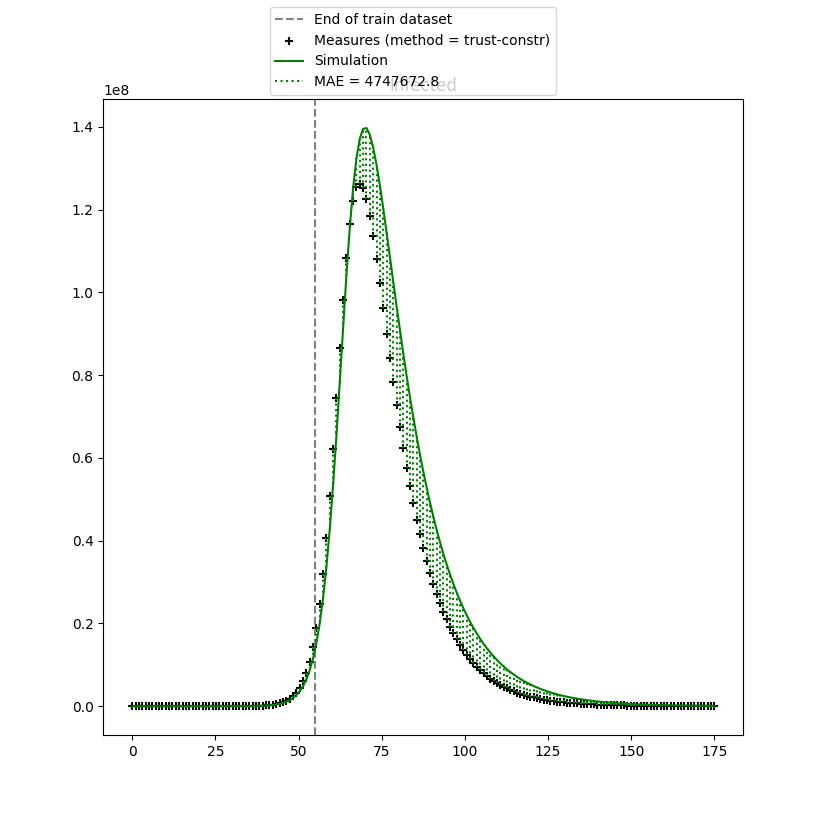}
        \caption{ SIRD }
        \label{fig:top3-sird-lowdata-nonoise-nosubgroups}
    \end{subfigure} 

    \begin{subfigure}[b]{\textwidth}
        \centering
        \includegraphics[width=0.3\linewidth, bb= 0 0 826 826]{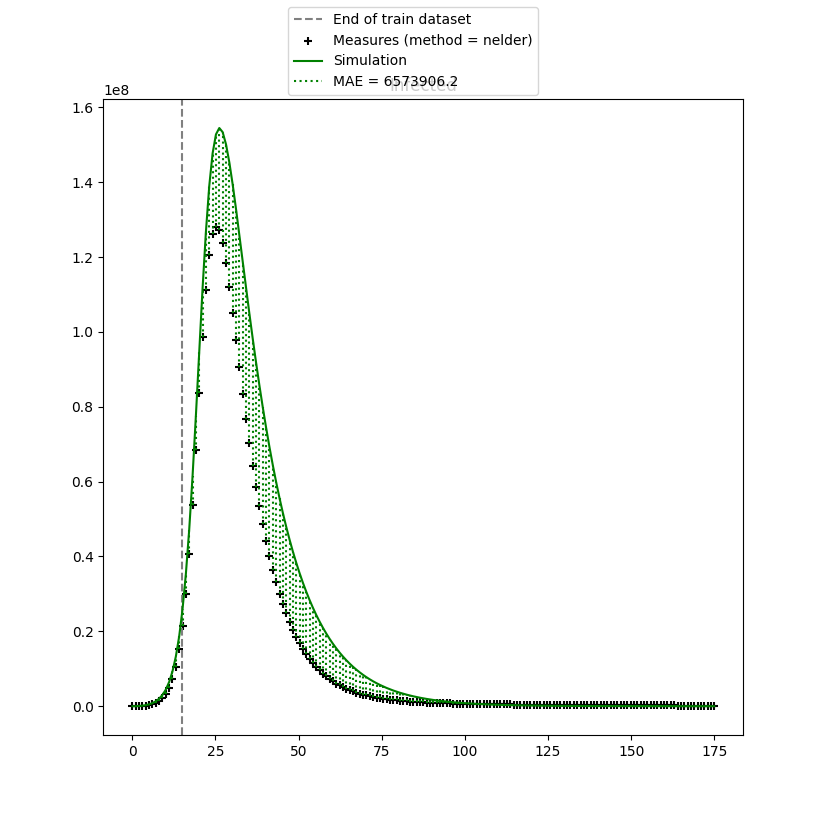}
        \includegraphics[width=0.3\linewidth, bb= 0 0 826 826]{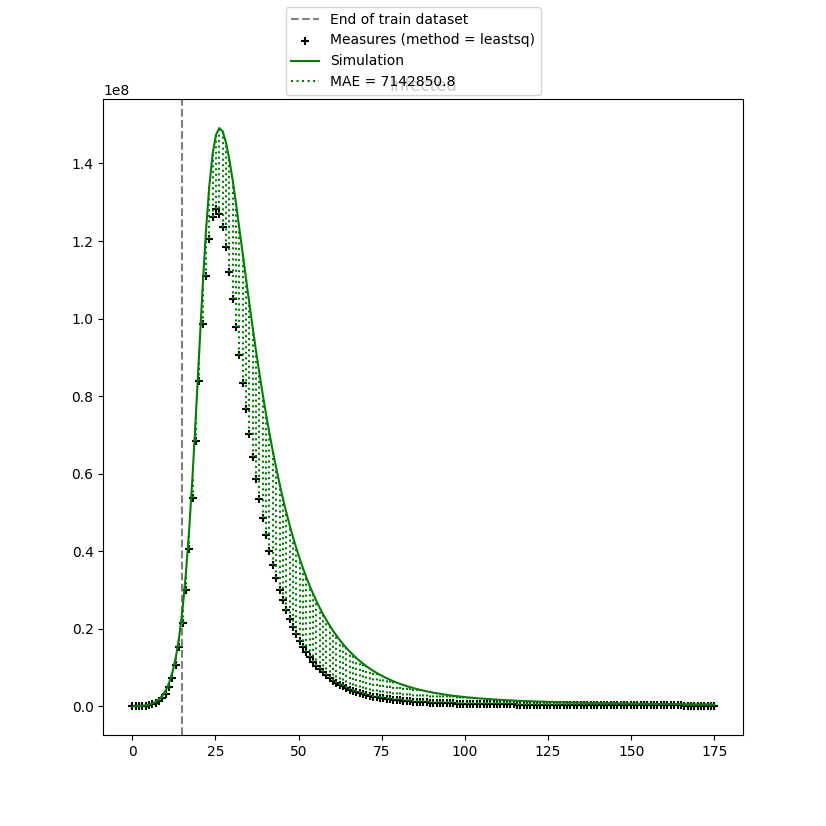}
        \includegraphics[width=0.3\linewidth, bb= 0 0 826 826]{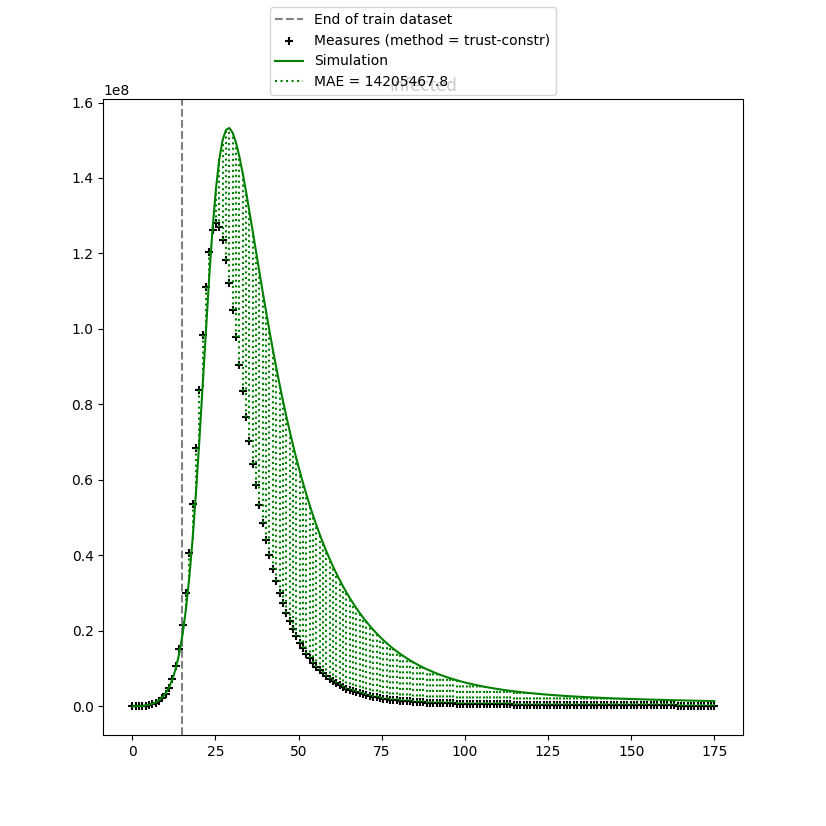}
        \caption{ SIRVD }
        \label{fig:top3-sirvd-lowdata-nonoise-nosubgroups}
    \end{subfigure} 

\caption{Best performing (top 3) calibration methods for all models (SIR, SIRD, and SIRVD) in the low data regime. 
}
\label{fig:top3-lowdata-nonoise-nosubgroups}
\end{figure}

\begin{figure}
\centering
    \begin{subfigure}[b]{\textwidth}
        \centering
        \includegraphics[width=0.3\linewidth, bb= 0 0 826 826]{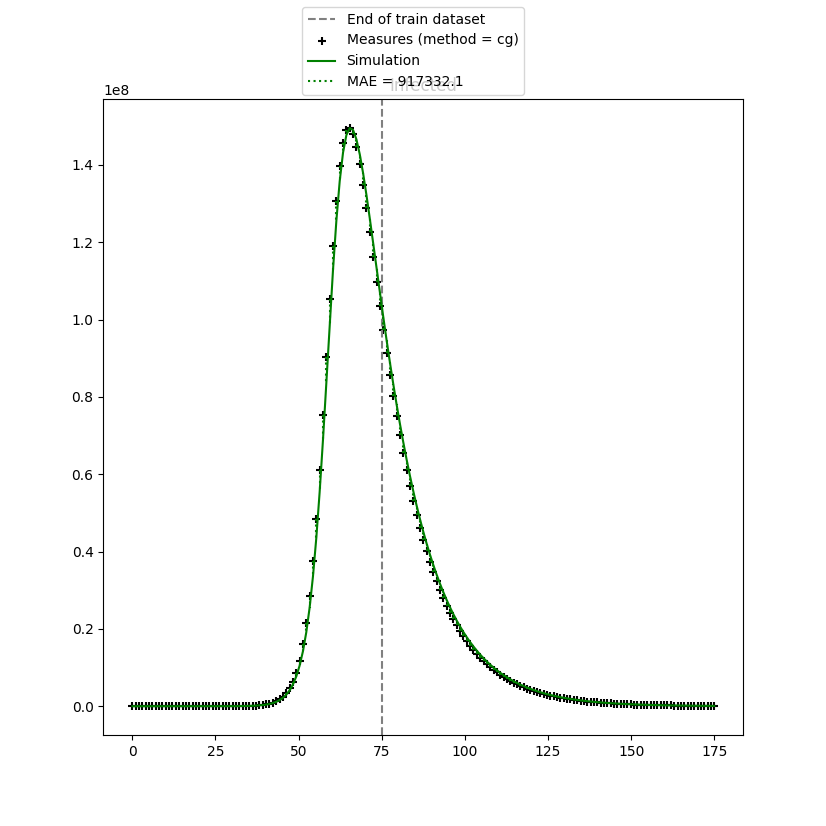}
        \includegraphics[width=0.3\linewidth, bb= 0 0 826 826]{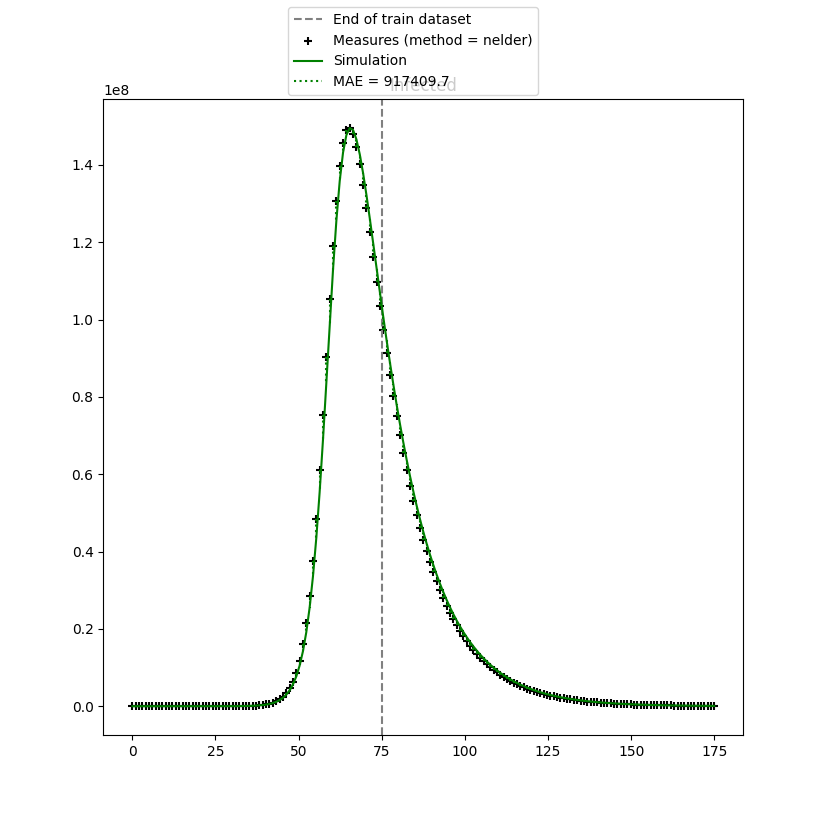}
        \includegraphics[width=0.3\linewidth, bb= 0 0 826 826]{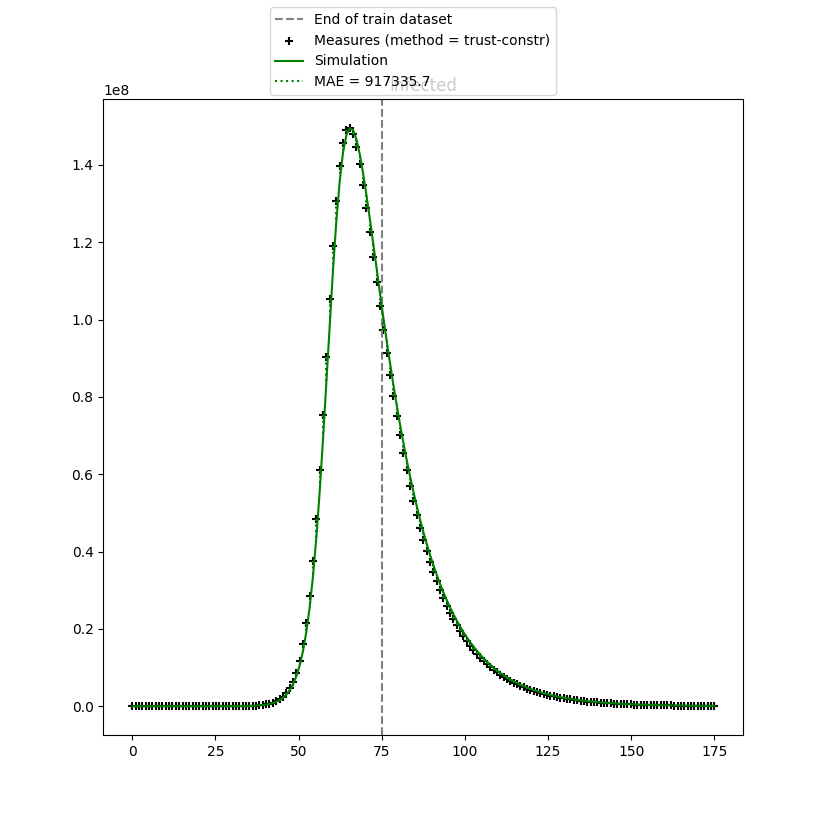}
        \caption{ SIR }
        \label{fig:top3-sir-highdata-nonoise-nosubgroups}
    \end{subfigure} 

    \begin{subfigure}[b]{\textwidth}
        \centering
        \includegraphics[width=0.3\linewidth, bb= 0 0 826 826]{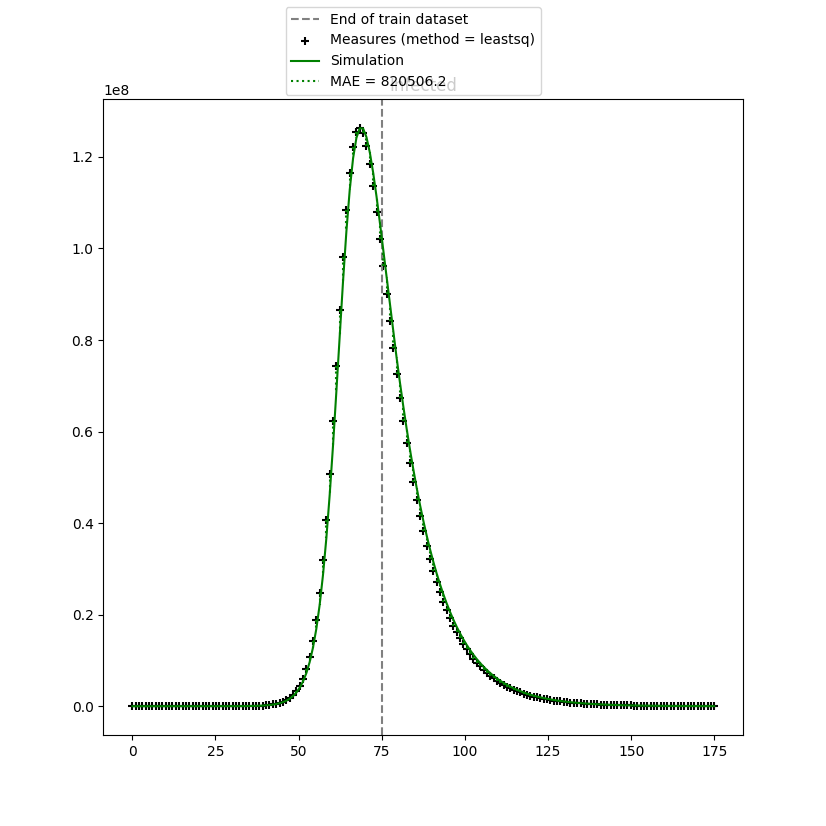}
        \includegraphics[width=0.3\linewidth, bb= 0 0 826 826]{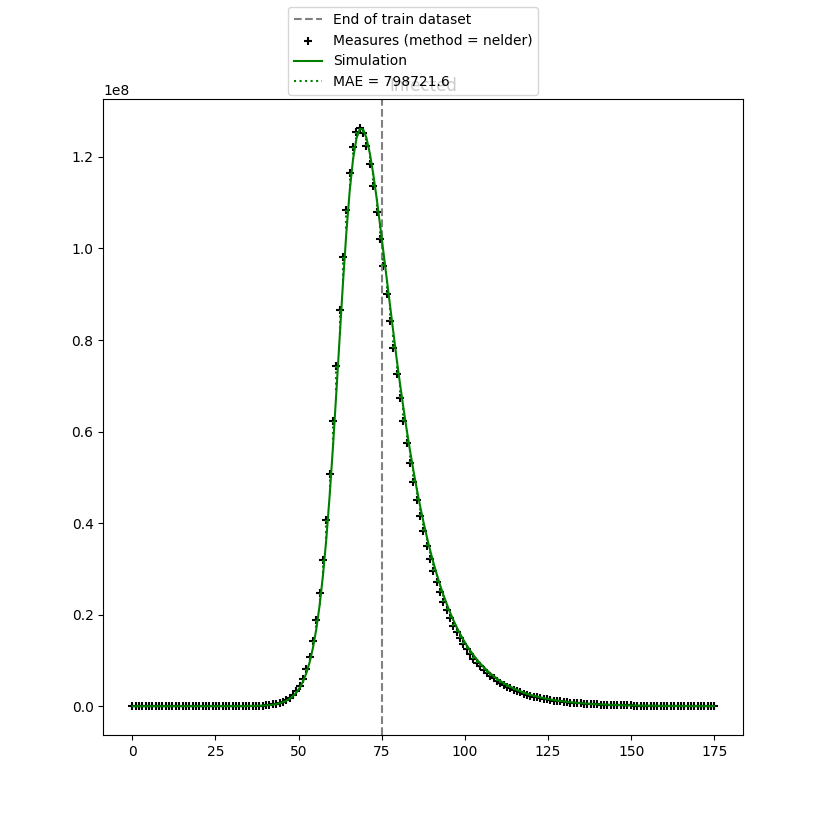}
        \includegraphics[width=0.3\linewidth, bb= 0 0 826 826]{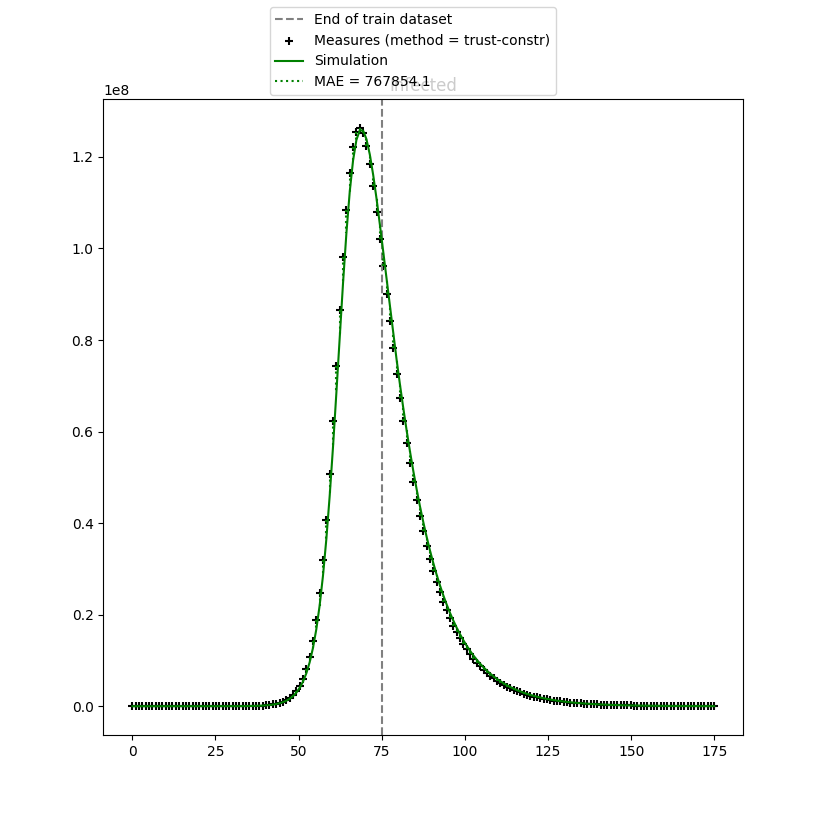}
        \caption{ SIRD }
        \label{fig:top3-sird-highdata-nonoise-nosubgroups}
    \end{subfigure} 

    \begin{subfigure}[b]{\textwidth}
        \centering
        \includegraphics[width=0.3\linewidth, bb= 0 0 826 826]{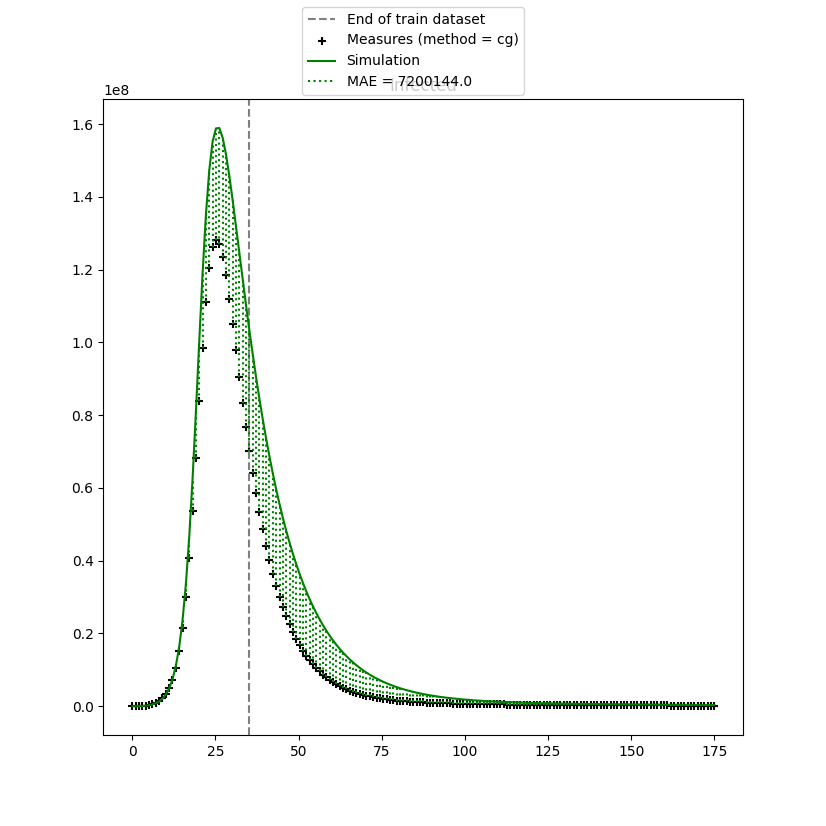}
        \includegraphics[width=0.3\linewidth, bb= 0 0 826 826]{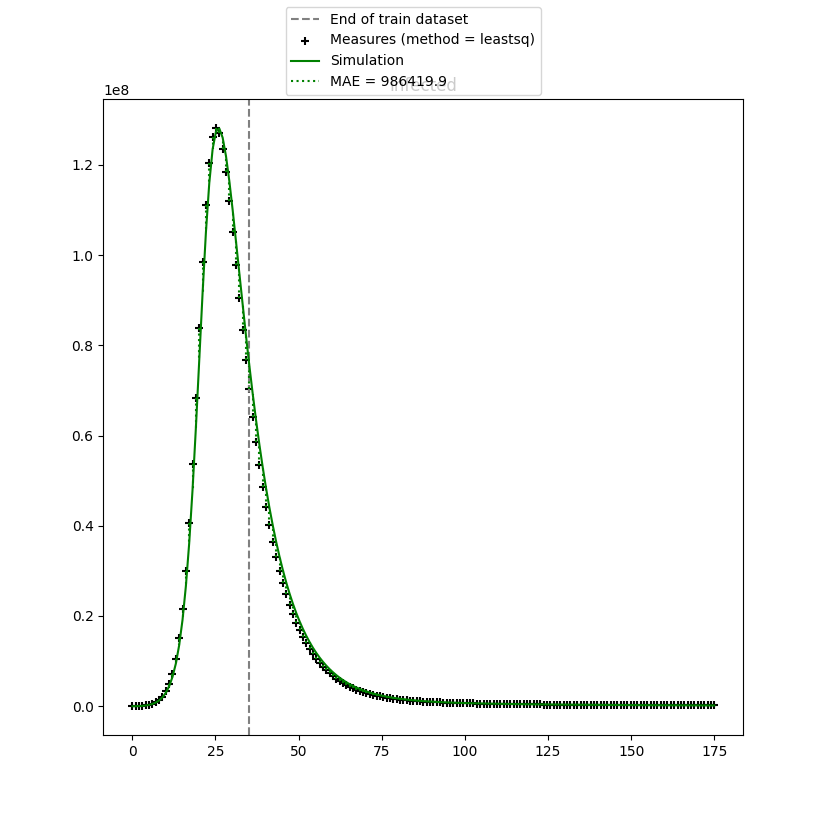}
        \includegraphics[width=0.3\linewidth, bb= 0 0 826 826]{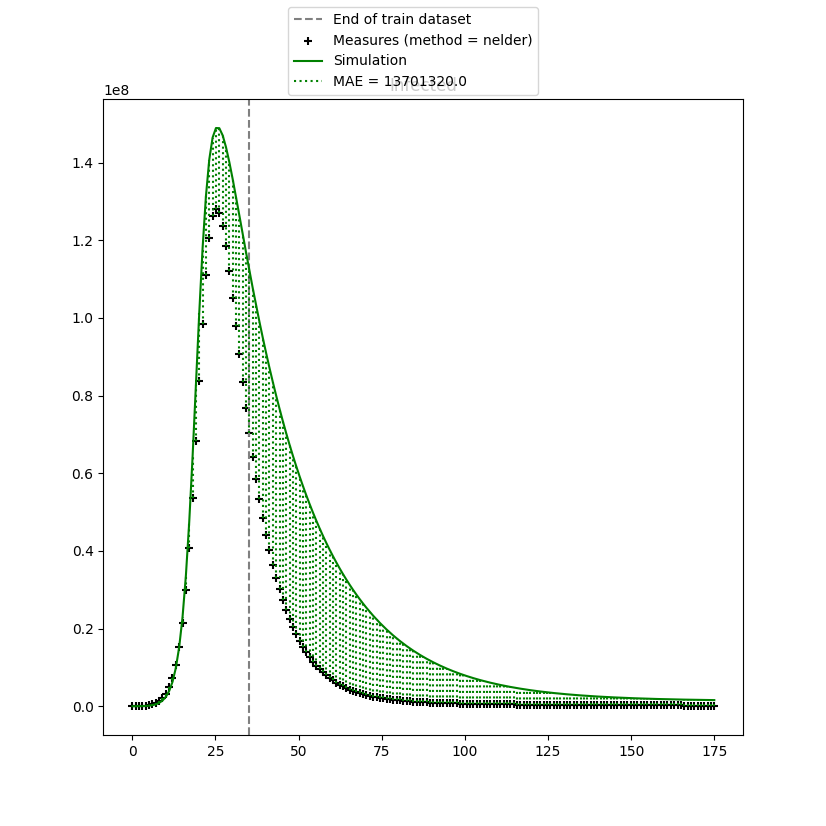}
        \caption{ SIRVD }
        \label{fig:top3-sirvd-highdata-nonoise-nosubgroups}
    \end{subfigure} 

\caption{Best performing (top 3) calibration methods for all models (SIR, SIRD, and SIRVD) in the high data regime.}
\label{fig:top3-highdata-nonoise-nosubgroups}
\end{figure}

We display the plots corresponding to the top-3 best-performing optimization methods on SIR, SIRD, and SIRVD in Figure ~\ref{fig:top3-lowdata-nonoise-nosubgroups}. The plots represent the curve of the number of \textit{Infected} people over time through the epidemic timeline (175 days chosen for our experiments). The black crosses ('+') represent the true number of infected people on each day and the green curve is obtained using the estimates of model parameters using the corresponding optimization method. The vertical black dotted line represents the amount of data (number of days) made available for training (the data points on the left of the line were used for training). The vertical green dotted lines represent the error between the predicted and the true value and we display the overall MAE at the headline of the plots. We define the number of days before the peak as the \textit{low-data regime} and the one after the peak as \textit{the high-data} regime. Intuitively, it is desirable to be able to work in the low-data regime in order to manage the epidemic before it gets very bad. However, that may not always be  feasible. Figure ~\ref{fig:top3-lowdata-nonoise-nosubgroups} compares the performances of all methods in the low data regime and displays the plots corresponding to the top 3 methods for each model --- SIR, SIRD, and SIRVD. No noise was assumed in this situation and no population subgroups were considered either. Figure \ref{fig:top3-highdata-nonoise-nosubgroups} shows the same experiments and results for the high data regime. 

Nelder-mead, Powell's method, and least squares minimization perform the best in SIR in low data regime, however, Nelder-mead did not make it in the top-3 in SIRD. Powell's method, least squares, and trust-region-based constrained minimization take the top 3 performance places in SIRD. Whereas, in SIRVD, Nelder-mead, least squares, and trust-region-based constrained minimization perform the best. It is clear that with changing model complexity, the best or most appropriate algorithm to choose becomes difficult. Nevertheless, all in all, Nelder-mead, Powell's method, least squares minimization, and trust-region-based constrained minimization empirically seem to be a nice short-list of calibration methods that can be used in this use-case (Figure ~\ref{fig:top3-lowdata-nonoise-nosubgroups}). On the other hand, we note that in the high data regime, Nelder-mead performs consistently among the top-3 on varying model complexities SIR, SIRD, and SIRVD. The other highlight algorithms are CG, least squares, and trust-region-based constrained minimization (Figure ~\ref{fig:top3-highdata-nonoise-nosubgroups}).

%% file: chapters/03_noise_in_data.tex
Noise in epidemic compartmental model data refers to random or systematic errors, uncertainties, or variability that can affect the accuracy or reliability of the data used in the model. This can arise from various sources, such as measurement errors, sampling errors, reporting errors, data entry errors, data collection biases, and other sources of variability in real-world data.

Noisy epidemic data can have significant implications on the model's outputs and conclusions. It can introduce uncertainties in the estimated parameters, affect the accuracy of predictions, and influence the reliability of policy recommendations or interventions. Therefore, it is important to account for and mitigate the impact of noise in data when calibrating, validating, and interpreting epidemic compartmental models. 

In this chapter, we add noise to previously simulated data and again calibrate the model parameters using all optimization methods to analyze and compare their performances across the amount of training data, model complexity, and noise levels. 

\subsection{Description of noisy simulated data} 

Adding noise to simulated data can be a useful way to make it more realistic and account for the inherent variability and uncertainty in real-world data. We added Gaussian noise to the simulated epidemiological compartmental models in the following steps: 

\begin{enumerate}

    \item Simulating the compartmental model without any noise, to obtain the true simulation data (as discussed in Section \ref{sec:simulated_data}). 

    \item Choosing an appropriate standard deviation for a Gaussian distribution that reflects the level of variability in the data to determine the level of noise to be added. 

    \item Generating a random sample of Gaussian noise with a mean of 0 and a standard deviation chosen in step 2. 

    \item Adding the noise to the simulated data at all time points by adding the noise to each data point in the simulation: 
        \begin{center}
               \textit{$data_{noisy}$} = \textit{$data_{true}$} + \textit{noise} 
        \end{center}
       where \textit{$data_{true}$} is the original simulated data, \textit{noise} is the generated Gaussian noise, and \textit{$data_{noisy}$} is the data with added noise. 

\end{enumerate} 

Gaussian noise is frequently used in epidemiological models (for instance, \cite{zhang2020dynamics,gu2011modeling}).
It is often a fair approximation of noise in real-world data, and being mathematically simple and versatile, it is appropriate for a wide range of modeling and statistical analyses. While other types of noise may be better suited to specific types of data, for these reasons, we choose Gaussian noise to model variability and uncertainty in our simulated epidemiological models. 

\subsection{Experiments and empirical results} 

\begin{figure}
\centering
    \begin{subfigure}[b]{\textwidth}
        \centering
        \includegraphics[width=0.3\linewidth, bb= 0 0 826 826]{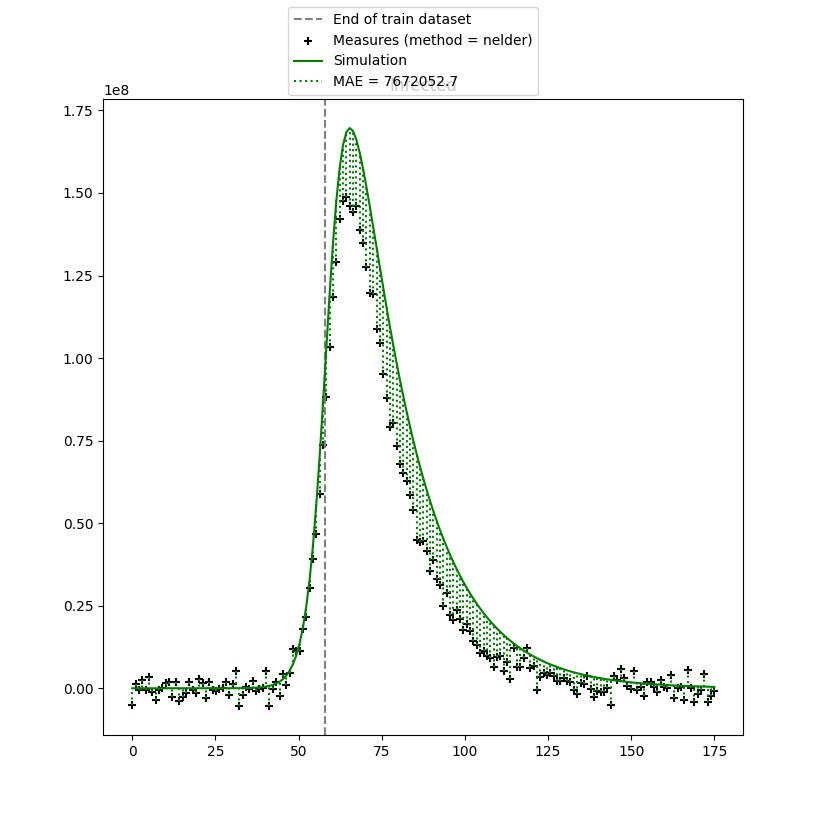}
        \includegraphics[width=0.3\linewidth, bb= 0 0 826 826]{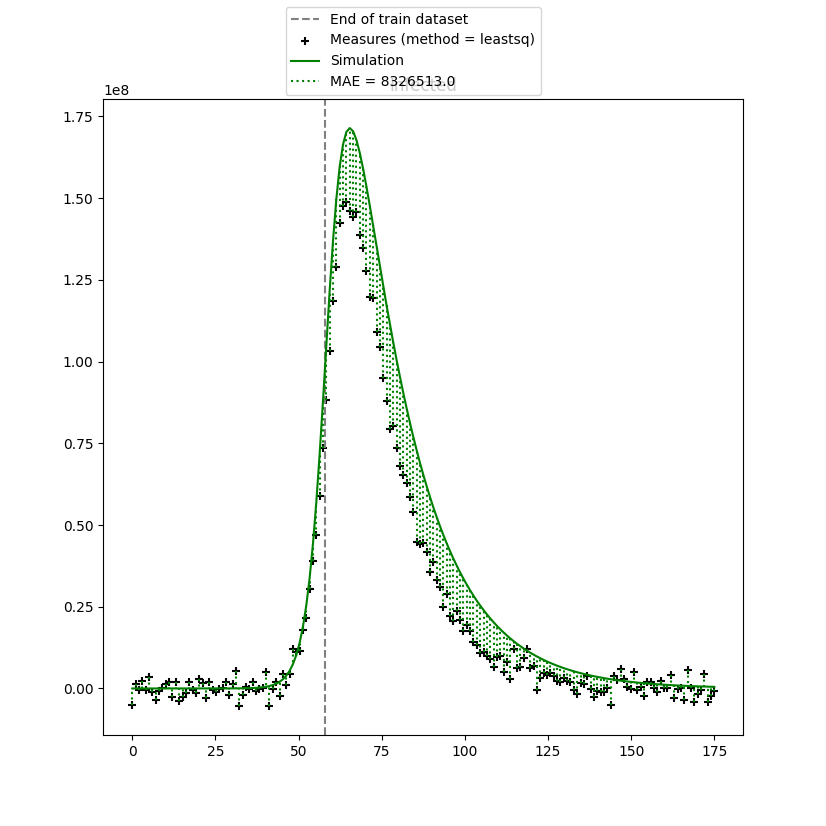}
        \includegraphics[width=0.3\linewidth, bb= 0 0 826 826]{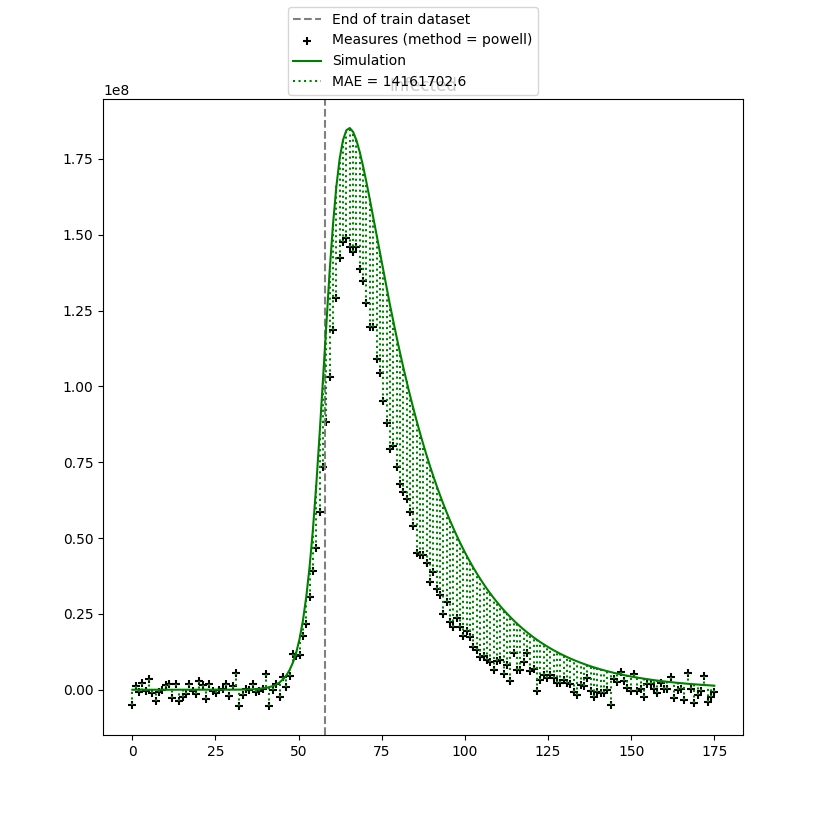}
        \caption{ SIR }
        \label{fig:top3-sir-lowdata-withnoise-nosubgroups}
    \end{subfigure} 

    \begin{subfigure}[b]{\textwidth}
        \centering
        \includegraphics[width=0.3\linewidth, bb= 0 0 826 826]{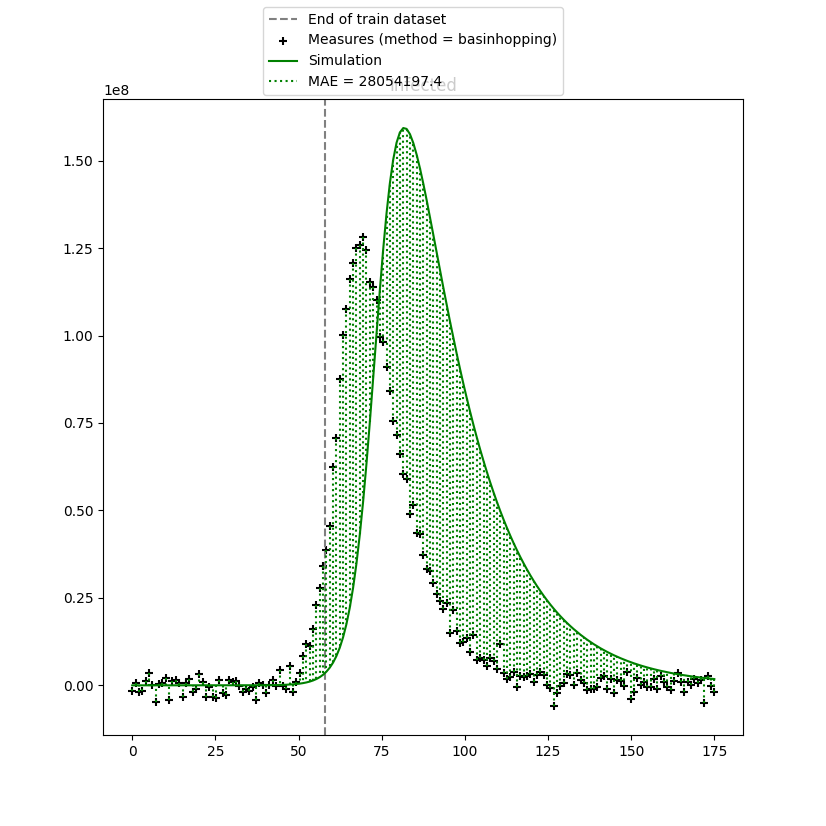}
        \includegraphics[width=0.3\linewidth, bb= 0 0 826 826]{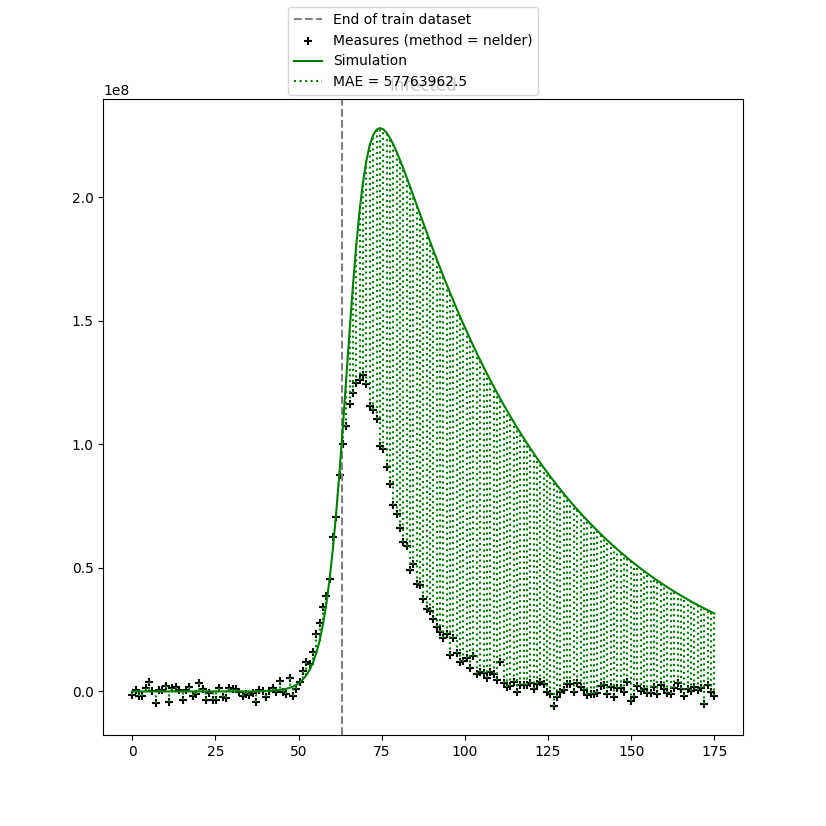}
        \includegraphics[width=0.3\linewidth, bb= 0 0 826 826]{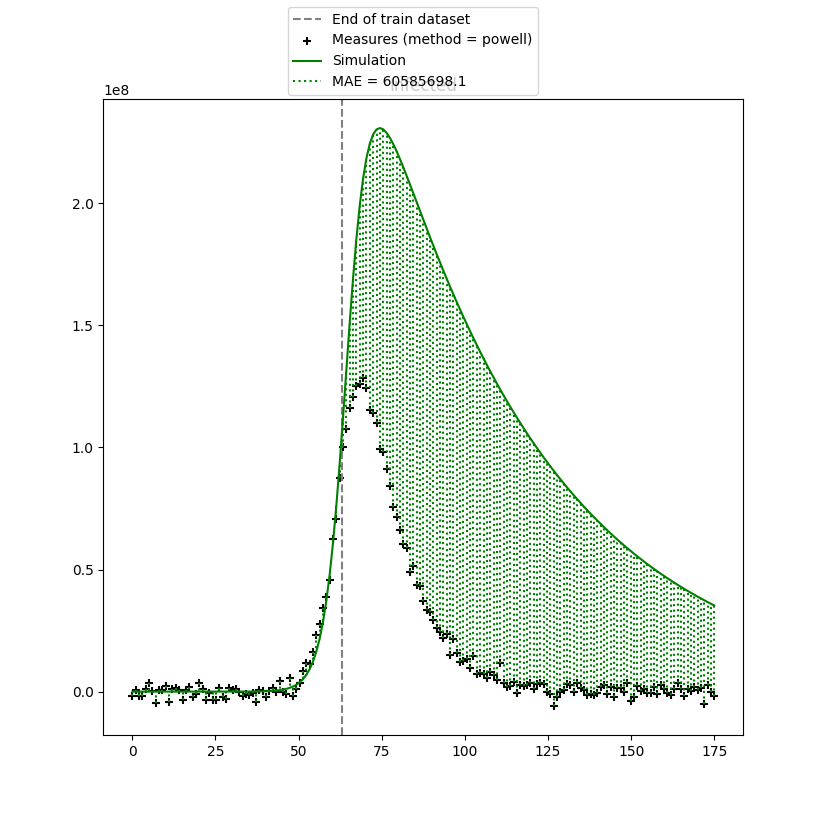}
        \caption{ SIRD }
        \label{fig:top3-sird-lowdata-withnoise-nosubgroups}
    \end{subfigure} 

    \begin{subfigure}[b]{\textwidth}
        \centering
        \includegraphics[width=0.3\linewidth, bb= 0 0 826 826]{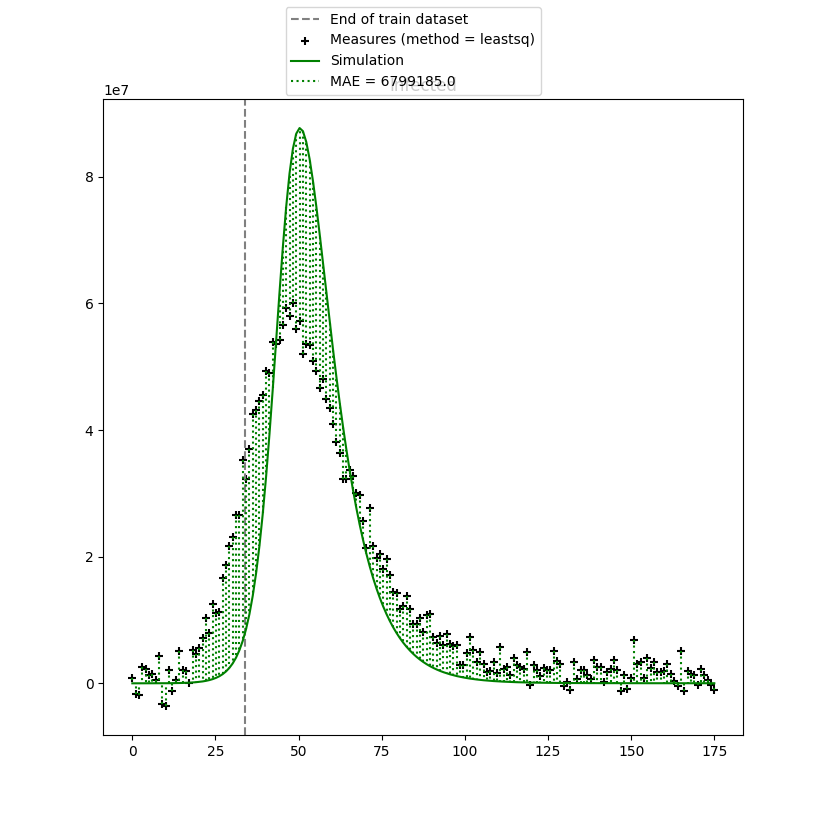}
        \includegraphics[width=0.3\linewidth, bb= 0 0 826 826]{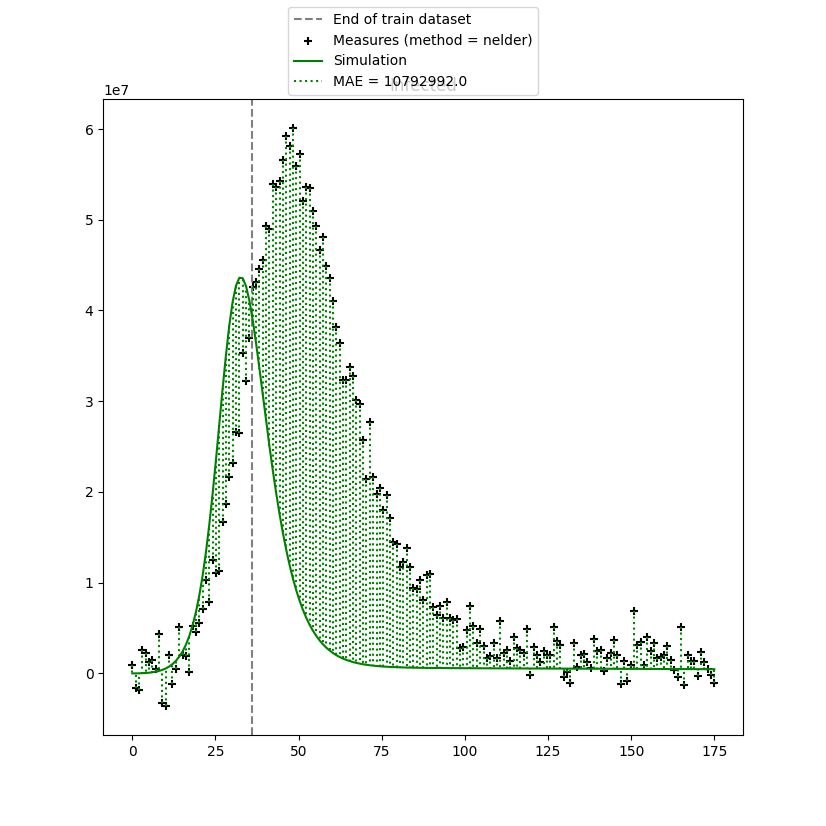}
        \includegraphics[width=0.3\linewidth, bb= 0 0 826 826]{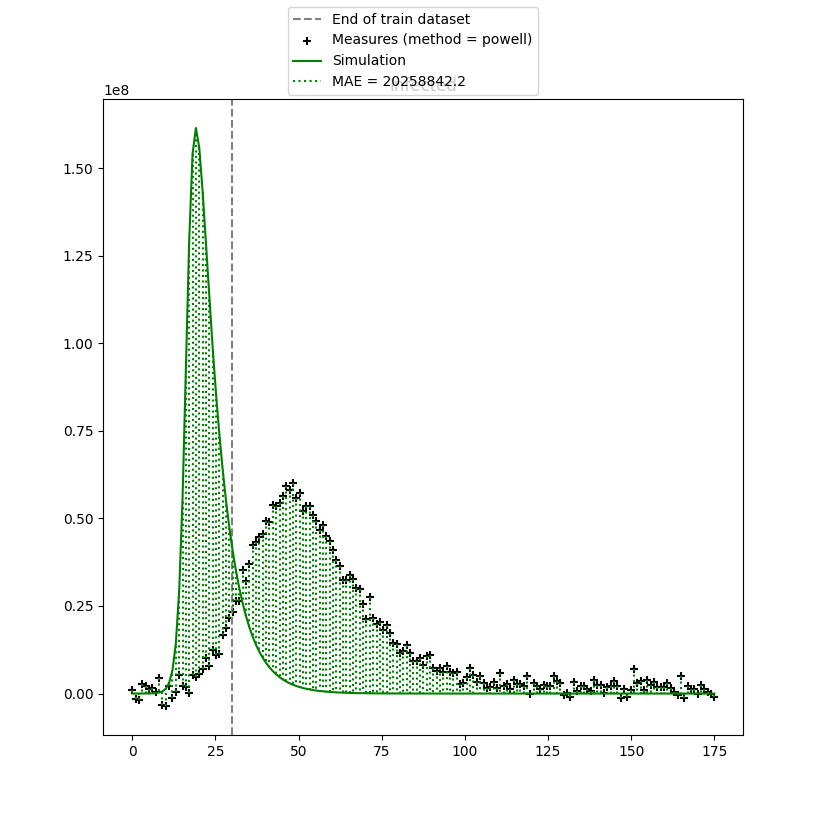}
        \caption{ SIRVD }
        \label{fig:top3-sirvd-lowdata-withnoise-nosubgroups}
    \end{subfigure} 

\caption{Best performing (top 3) calibration methods for all models (SIR, SIRD, and SIRVD) with noisy data in the low data regime.}
\label{fig:top3-lowdata-withnoise-nosubgroups}
\end{figure}

Figure \ref{fig:top3-lowdata-withnoise-nosubgroups} shows our results of the top three performing calibration (optimization) methods trained on a low data regime, for all model complexities (SIR, SIRD, and SIRVD) with Gaussian noise added to the data. In SIR, in the low data regime (corresponding to 58 days here), Nelder-mead, Powell's method, and least squares minimization perform the best. In SIRD (with 62 days in the low data regime), Nelder-mead and Powell's method still make it to the top, however, basinhopping method performs better than least squares minimization. We note here that the difference in MAE for the basinhopping method and least squares optimization is very low which means the basinhopping method performed only slightly better. On further increasing the model complexity to SIRVD (with 26 days in the low data regime), the top 3 performances are again obtained Nelder-mead, Powell's method, and least squares minimization. So, in the case of noisy simulated data, through our experiments, it can be safe to say that Nelder-mead, Powell's method, and least squares minimization perform reasonably better than all the other evaluated methods. Nevertheless, we note here that, the performance of even these top methods is very good as can be seen in Figure \ref{fig:top3-lowdata-withnoise-nosubgroups}. The MAE values in all the curves, especially with increasing model complexity, yield results that are not practical. 

\begin{figure}
\centering
    \begin{subfigure}[b]{\textwidth}
        \centering
        \includegraphics[width=0.3\linewidth, bb= 0 0 826 826]{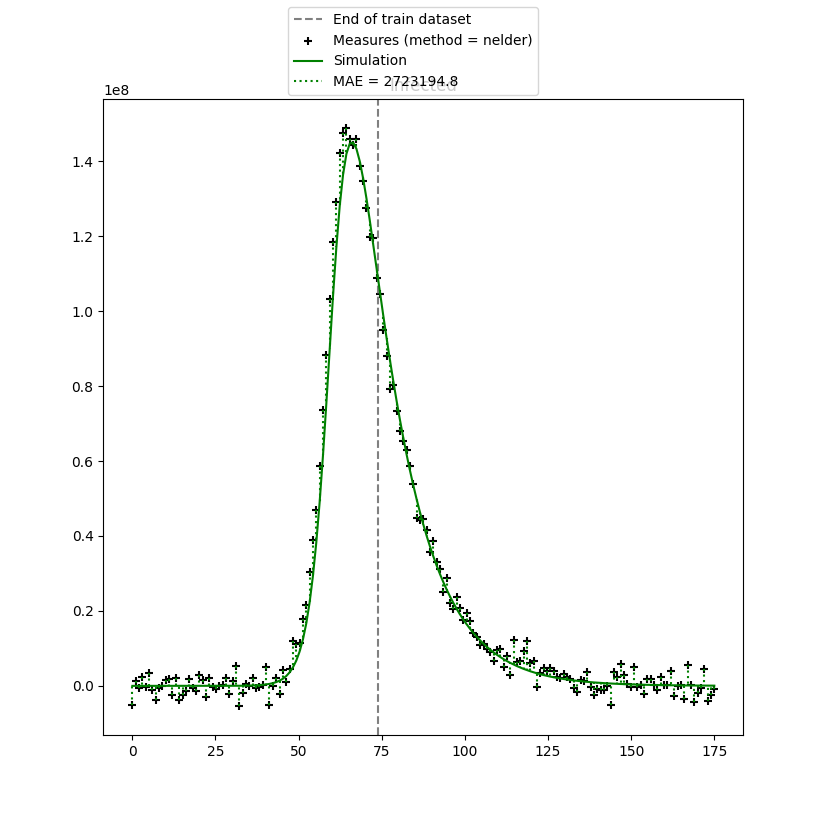}
        \includegraphics[width=0.3\linewidth, bb= 0 0 826 826]{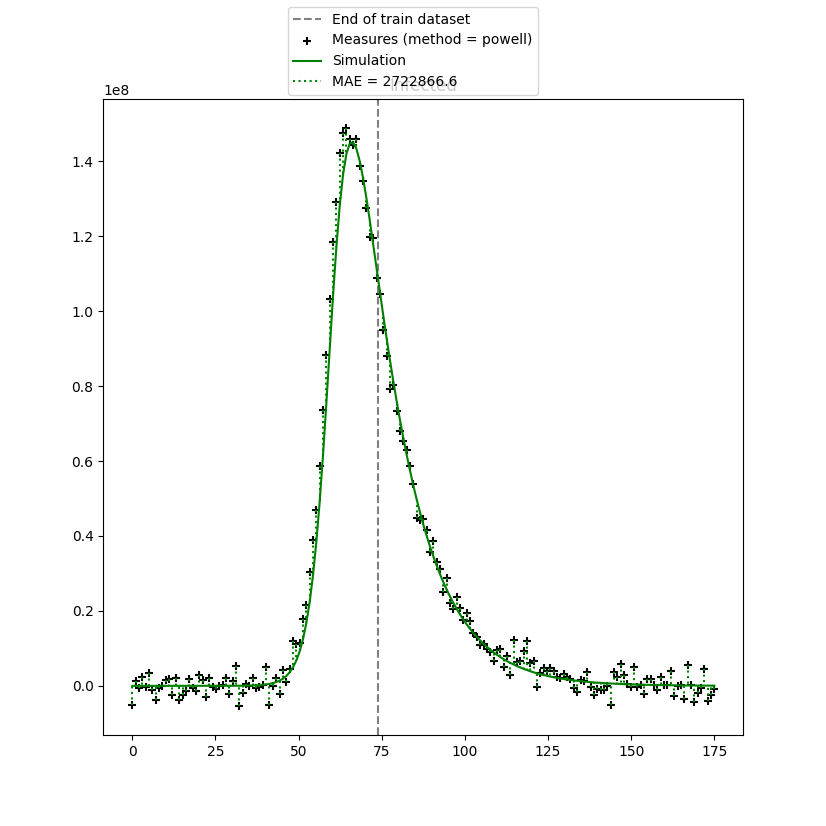}
        \includegraphics[width=0.3\linewidth, bb= 0 0 826 826]{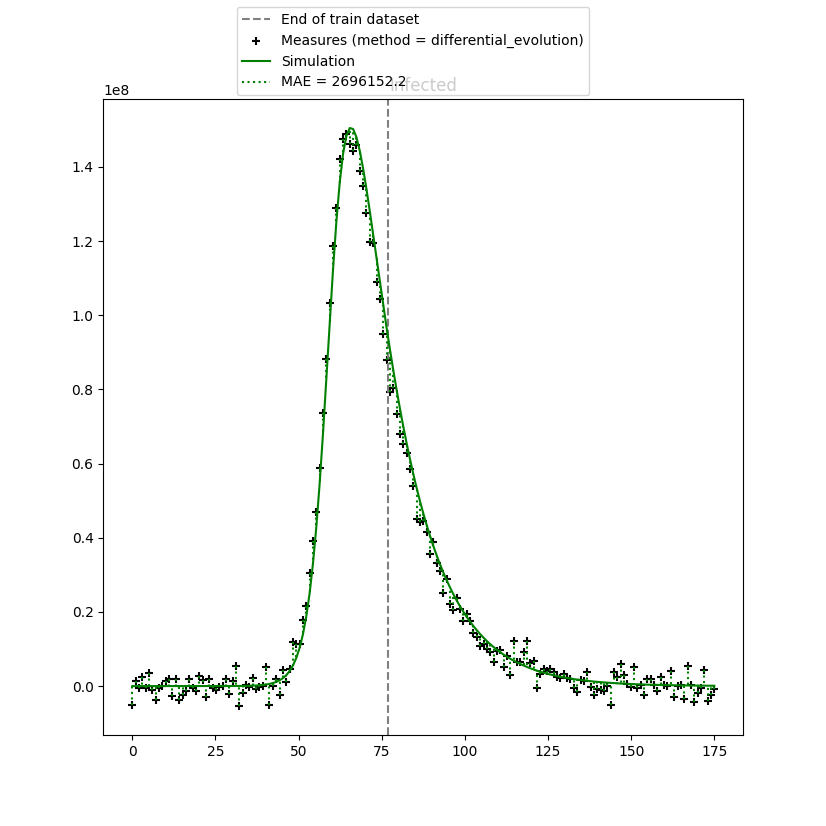}
        \caption{ SIR }
        \label{fig:top3-sir-highdata-withnoise-nosubgroups}
    \end{subfigure} 

    \begin{subfigure}[b]{\textwidth}
        \centering
        \includegraphics[width=0.3\linewidth, bb= 0 0 826 826]{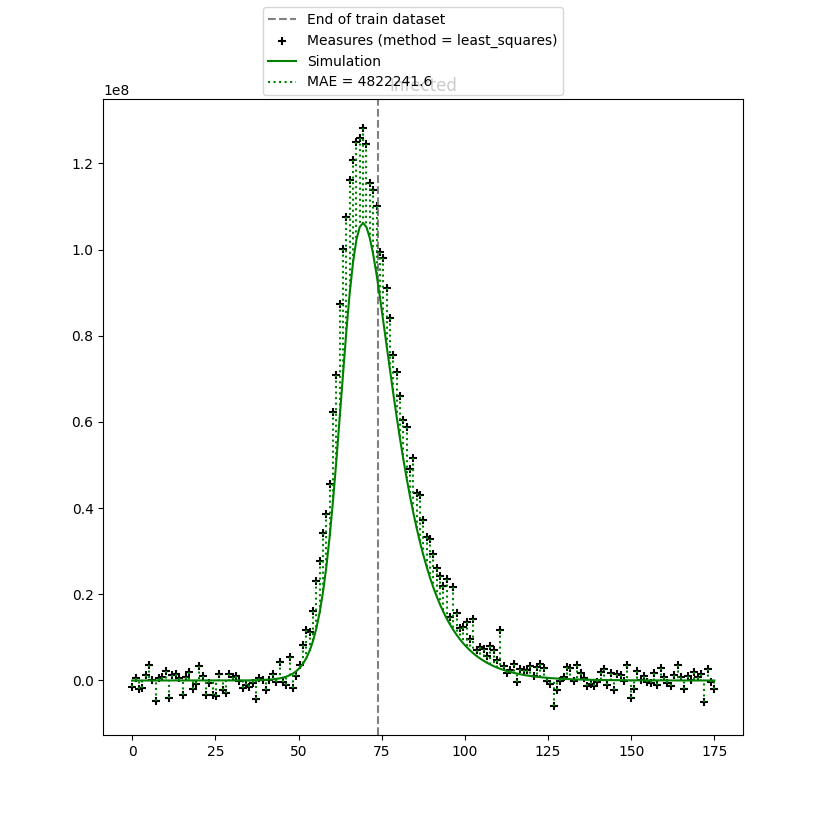}
        \includegraphics[width=0.3\linewidth, bb= 0 0 826 826]{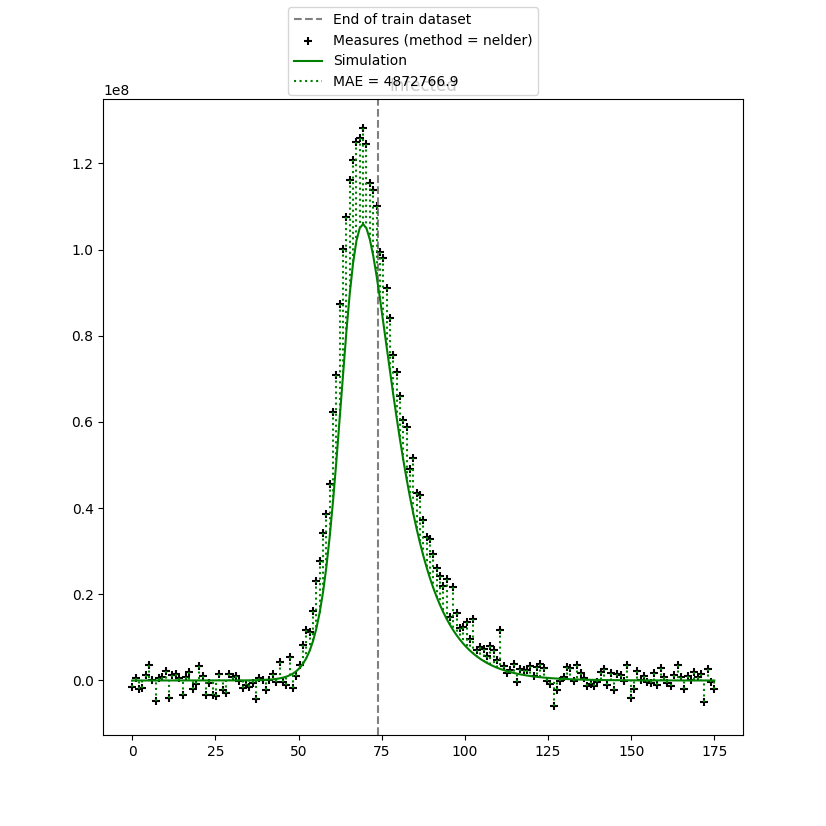}
        \includegraphics[width=0.3\linewidth, bb= 0 0 826 826]{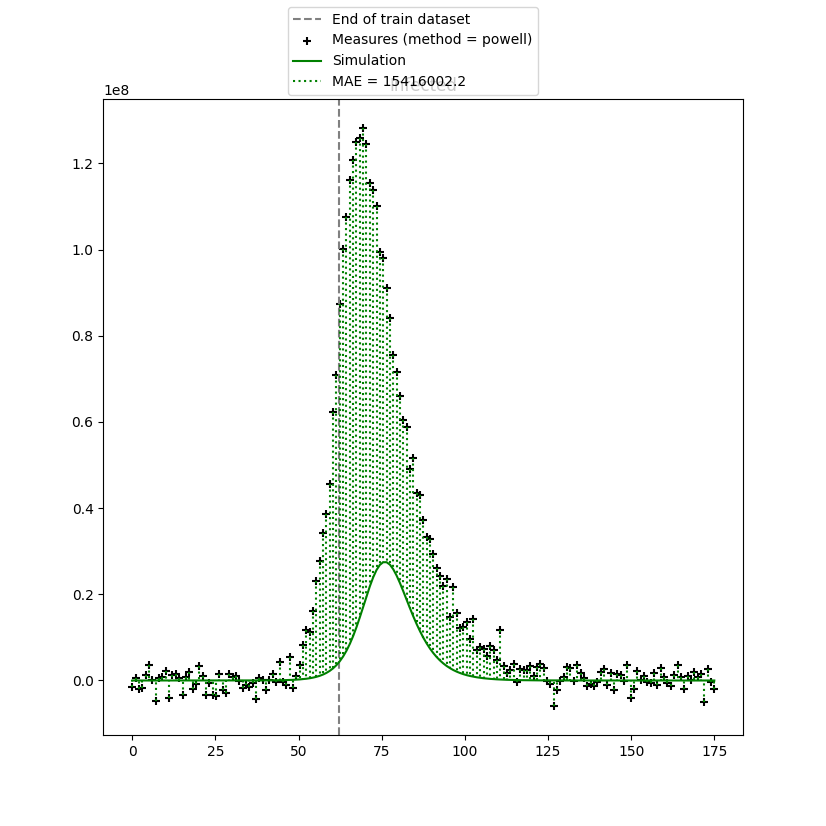}
        \caption{ SIRD }
        \label{fig:top3-sird-highdata-withnoise-nosubgroups}
    \end{subfigure} 

    \begin{subfigure}[b]{\textwidth}
        \centering
        \includegraphics[width=0.3\linewidth, bb= 0 0 826 826]{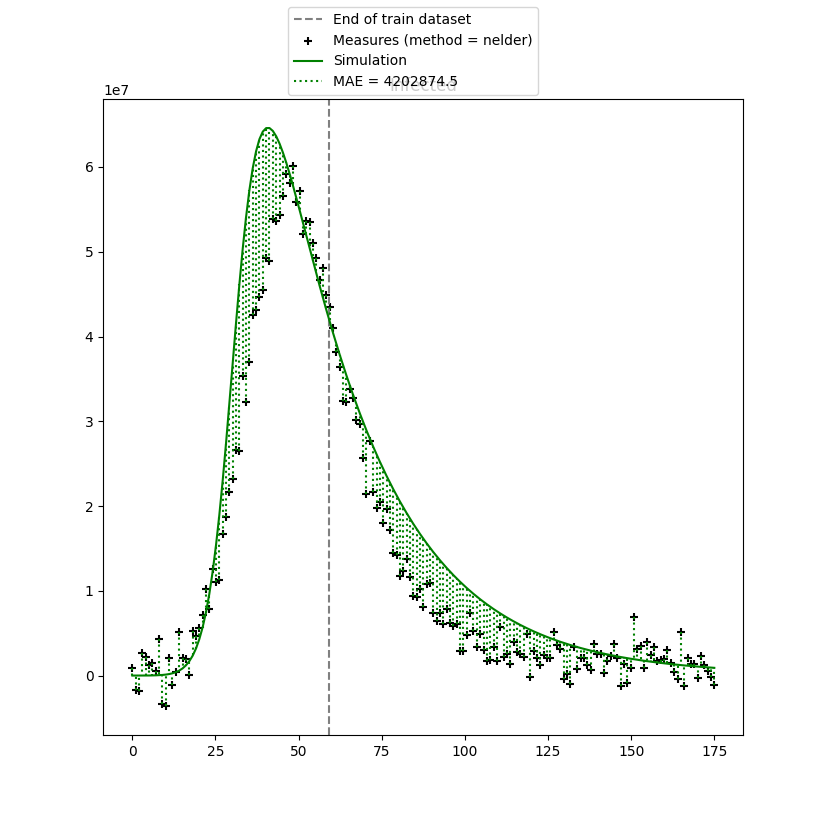}
        \includegraphics[width=0.3\linewidth, bb= 0 0 826 826]{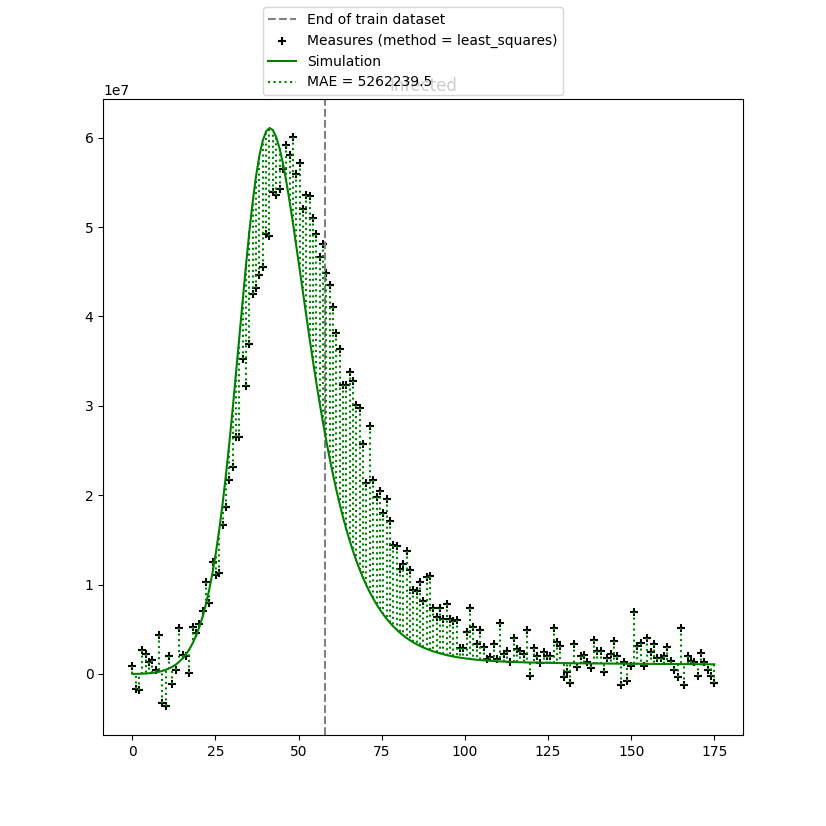}
        \includegraphics[width=0.3\linewidth, bb= 0 0 826 826]{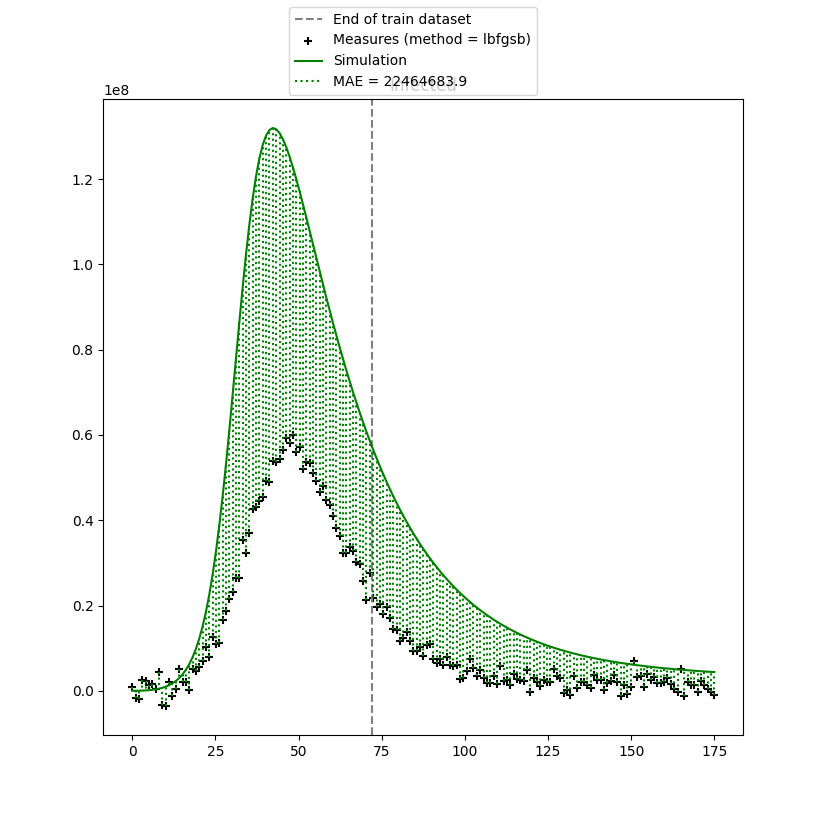}
        \caption{ SIRVD }
        \label{fig:top3-sirvd-highdata-withnoise-nosubgroups}
    \end{subfigure} 

\caption{Best performing (top 3) calibration methods for all models (SIR, SIRD, and SIRVD) with noisy data in the high data regime.} 
\label{fig:top3-highdata-withnoise-nosubgroups}
\end{figure}

Dealing with noise is a difficult and challenging problem and requires further work. Figure \ref{fig:top3-highdata-withnoise-nosubgroups} shows our results for the high data regime. The Nelder-Mead method and least squares minimization still dominate the results across model complexities. However, DE and L-BFGS-B end up performing better than Powell's method in SIR and SIRVD. We note here that predictions made after the peak in the \textit{Infected} curve are of limited utility because the epidemic is already waning. Moreover, many methods can perform better when given more data. 

%% file: chapters/04_population_subgroups.tex
Epidemic compartmental models with \textbf{\textit{population subgroups}} and a \textbf{\textit{contact matrix}} are a type of mathematical modeling approach used to study the spread of infectious diseases within a population that is divided into different subgroups while taking into account the contact patterns between these subgroups. This allows for a more detailed and realistic representation of disease spread within a population, because it accounts for the heterogeneity in contact patterns and disease dynamics among different subpopulations. 

When \textit{subgroups} are considered, the population is separated into compartments and  further separated based on relevant criteria such as age, geographical location, occupation, or risk category. Each subgroup has its own set of compartments that indicate the subgroup's sickness status. For example, if age is considered as a subgroup, there would be separate compartments for susceptible, infected, recovered, vaccinated, and deceased individuals in each age group. 

The \textit{contact matrix} represents the frequency and intensity of interactions or contacts between different subgroups. It is typically represented as a square matrix, where the entries denote the rate or probability of contacts between pairs of subgroups. The contact matrix can be estimated from empirical data, such as surveys or social contact studies, or it can be derived from assumptions about the contact patterns between subgroups based on expert knowledge or other relevant information.

Calibration of epidemic compartmental models with subgroups and a contact matrix involves estimating the model parameters, including the transmission rates, recovery rates, and mortality rates, separately for each subgroup. This may require fitting the model to subgroup-specific data, incorporating the contact matrix into the model, and adjusting the parameters to minimize the difference between observed data and model predictions for each subgroup. 

Epidemic compartmental models with population subgroups and contact matrices can be a valuable tool, however, such modeling also comes with its own challenges, such as the availability and quality of subgroup-specific data, the complexity of estimating parameters for multiple subgroups, and the need for careful validation and refinement of the calibrated model. 

\subsection{Generating Epidemiological Data with Population Subgroups using EpiPolicy}  

Simulating epidemiological data with population subgroups involves creating a hypothetical population and generating data on health outcomes and risk factors within specific subgroups. This approach can be useful for testing hypotheses, exploring different scenarios, and evaluating interventions aimed at reducing health disparities. 

We used EpiPolicy\footnote{\url{https://epipolicy.github.io/}} \cite{mai2022epipolicy,tariq2021planning} to generate epidemiological data with population subgroups. It is a population-based epidemic simulator and policy tool that lets users adjust its compartmental model to capture different epidemic situations for diseases such as COVID-19. Its user interface makes it simple to describe numerous interventions such as social distancing, school closure, immunization, and disease-specific therapies. EpiPolicy assists policymakers by modeling and building locale-specific intervention regimens that reduce disease burden while minimizing social and economic costs. 

In particular, we implemented the following steps: 

\begin{enumerate}
    
    \item Defined the \textbf{\textit{population}} of interest considering the total number of individuals with relevant characteristics such as age, gender, race or ethnicity, and socioeconomic position. 

    \item Identified the \textbf{\textit{subgroups}} of interest within the population, based on age. (Note: any characteristic from step 1 can be selected here). 

    \item Specified the \textbf{\textit{health outcomes}} and \textbf{\textit{risk factors}} of interest for each subgroup, mainly including disease incidence, prevalence, mortality, and relevant risk factors, such as physical activity. (This is done by defining the contact matrix for the population subgroups in Epipolicy). 

    \item Used EpiPolicy to \textbf{\textit{generate data}} on the specifications made for each subgroup from the previous steps. 

\end{enumerate} 

Although simulating epidemiological data with population subgroups and contact matrices can be a powerful tool for generating insights into the complex interplay between health outcomes and risk factors in different subgroups, it is important to note that any simulation model is only as good as the assumptions and parameters used to build it, and the results of simulations should be interpreted with caution. 

\subsection{Discussion of experiments and results} 

We considered age-based population subgroups --- children, adults, and seniors, and defined the contact matrix among population subgroups as part of numerous facilities. A \textit{facility} is any area in a location where different population groups spend time and engage with one another. Many facilities, such as households, schools, workplaces, community centers, places of worship, shopping malls, and many more, may be of interest in the context of epidemic modeling. We described two types of data: 

\begin{enumerate}

    \item The proportion of time that individuals in a group spend in any facility (e.g. Children spend ~35\% of their time in schools) --- Table~\ref{table:facilty_matrix}, and  
    
    \item The proportion of time that individuals in a group spend interacting with individuals in other groups in any facility (e.g. Children spend ~90\% of their time interacting with other children in schools and ~7\% of their time interacting with teachers) --- Table~\ref{table:contact_matrix}. 

\end{enumerate}

\begin{table}[ht]
\begin{tabular}{ccccc}
                                                        &                                         & \multicolumn{3}{c}{\textbf{Subgroup}}                                                                        \\ \cline{3-5} 
                                                        & \multicolumn{1}{c|}{\textbf{}}          & \multicolumn{1}{c|}{\textit{Children}} & \multicolumn{1}{c|}{\textit{Adults}} & \multicolumn{1}{c|}{Seniors} \\ \cline{2-5} 
\multicolumn{1}{c|}{\multirow{4}{*}{\textbf{Facility}}} & \multicolumn{1}{c|}{\textit{Household}} & \multicolumn{1}{c|}{0.4}               & \multicolumn{1}{c|}{0.4}             & \multicolumn{1}{c|}{0.54}    \\ \cline{2-5} 
\multicolumn{1}{c|}{}                                   & \multicolumn{1}{c|}{\textit{School}}    & \multicolumn{1}{c|}{0.31}              & \multicolumn{1}{c|}{0.08}            & \multicolumn{1}{c|}{0.01}    \\ \cline{2-5} 
\multicolumn{1}{c|}{}                                   & \multicolumn{1}{c|}{\textit{Workplace}} & \multicolumn{1}{c|}{0.08}              & \multicolumn{1}{c|}{0.32}            & \multicolumn{1}{c|}{0.18}    \\ \cline{2-5} 
\multicolumn{1}{c|}{}                                   & \multicolumn{1}{c|}{\textit{Community}} & \multicolumn{1}{c|}{0.2}               & \multicolumn{1}{c|}{0.2}             & \multicolumn{1}{c|}{0.27}    \\ \cline{2-5} 
\end{tabular}
\caption{Proportion of time spent by each population in each facility.}
\label{table:facilty_matrix}
\end{table}

\begin{table}[ht]
\begin{tabular}{ccccccc}
\textbf{}                                        &                                        &                                        & \textit{}                            & \textit{}                             &  &  \\ \cline{3-5}
                                                 & \multicolumn{1}{c|}{}                  & \multicolumn{1}{c|}{\textit{Children}} & \multicolumn{1}{c|}{\textit{Adults}} & \multicolumn{1}{c|}{\textit{Seniors}} &  &  \\ \cline{2-5}
\multicolumn{1}{c|}{\textit{\textbf{Household}}} & \multicolumn{1}{c|}{\textit{Children}} & \multicolumn{1}{c|}{0.37}              & \multicolumn{1}{c|}{0.53}            & \multicolumn{1}{c|}{0.1}              &  &  \\ \cline{2-5}
\multicolumn{1}{c|}{\textbf{}}                   & \multicolumn{1}{c|}{\textit{Adults}}   & \multicolumn{1}{c|}{0.32}              & \multicolumn{1}{c|}{0.6}             & \multicolumn{1}{c|}{0.07}             &  &  \\ \cline{2-5}
\multicolumn{1}{c|}{\textbf{}}                   & \multicolumn{1}{c|}{\textit{Seniors}}  & \multicolumn{1}{c|}{0.27}              & \multicolumn{1}{c|}{0.36}            & \multicolumn{1}{c|}{0.37}             &  &  \\ \cline{2-5}
\textbf{}                                        &                                        &                                        &                                      &                                       &  &  \\ \cline{2-5}
\multicolumn{1}{c|}{\textit{\textbf{School}}}    & \multicolumn{1}{c|}{\textit{Children}} & \multicolumn{1}{c|}{0.92}              & \multicolumn{1}{c|}{0.08}            & \multicolumn{1}{c|}{0}                &  &  \\ \cline{2-5}
\multicolumn{1}{c|}{\textit{\textbf{}}}          & \multicolumn{1}{c|}{\textit{Adults}}   & \multicolumn{1}{c|}{0.67}              & \multicolumn{1}{c|}{0.33}            & \multicolumn{1}{c|}{0}                &  &  \\ \cline{2-5}
\multicolumn{1}{c|}{\textit{\textbf{}}}          & \multicolumn{1}{c|}{\textit{Seniors}}  & \multicolumn{1}{c|}{0.75}              & \multicolumn{1}{c|}{0.25}            & \multicolumn{1}{c|}{0}                &  &  \\ \cline{2-5}
\textbf{}                                        &                                        &                                        &                                      &                                       &  &  \\ \cline{2-5}
\multicolumn{1}{c|}{\textit{\textbf{Workplace}}} & \multicolumn{1}{c|}{\textit{Children}} & \multicolumn{1}{c|}{0}                 & \multicolumn{1}{c|}{0.89}            & \multicolumn{1}{c|}{0.11}             &  &  \\ \cline{2-5}
\multicolumn{1}{c|}{\textit{\textbf{}}}          & \multicolumn{1}{c|}{\textit{Adults}}   & \multicolumn{1}{c|}{\textit{0.03}}     & \multicolumn{1}{c|}{0.92}            & \multicolumn{1}{c|}{0.05}             &  &  \\ \cline{2-5}
\multicolumn{1}{c|}{\textit{\textbf{}}}          & \multicolumn{1}{c|}{\textit{Seniors}}  & \multicolumn{1}{c|}{\textit{0.04}}     & \multicolumn{1}{c|}{0.94}            & \multicolumn{1}{c|}{0.02}             &  &  \\ \cline{2-5}
\textbf{}                                        &                                        &                                        &                                      &                                       &  &  \\ \cline{2-5}
\multicolumn{1}{c|}{\textit{\textbf{Community}}} & \multicolumn{1}{c|}{\textit{Children}} & \multicolumn{1}{c|}{0.54}              & \multicolumn{1}{c|}{0.4}             & \multicolumn{1}{c|}{0.06}             &  &  \\ \cline{2-5}
\multicolumn{1}{c|}{\textit{}}                   & \multicolumn{1}{c|}{\textit{Adults}}   & \multicolumn{1}{c|}{\textit{0.1}}      & \multicolumn{1}{c|}{0.57}            & \multicolumn{1}{c|}{0.34}             &  &  \\ \cline{2-5}
\multicolumn{1}{c|}{\textit{}}                   & \multicolumn{1}{c|}{\textit{Seniors}}  & \multicolumn{1}{c|}{\textit{0.1}}      & \multicolumn{1}{c|}{0.57}            & \multicolumn{1}{c|}{0.34}             &  &  \\ \cline{2-5}
\end{tabular}
\caption{Proportion of time that individuals in a group spend interacting with individuals in other groups in any facility.}
\label{table:contact_matrix}
\end{table}

\begin{figure}[ht]
\centering
    \begin{subfigure}[b]{\textwidth}
        \centering
        \begin{subfigure}[b]{0.3\textwidth}
            \includegraphics[width=\linewidth, bb= 0 0 826 826]{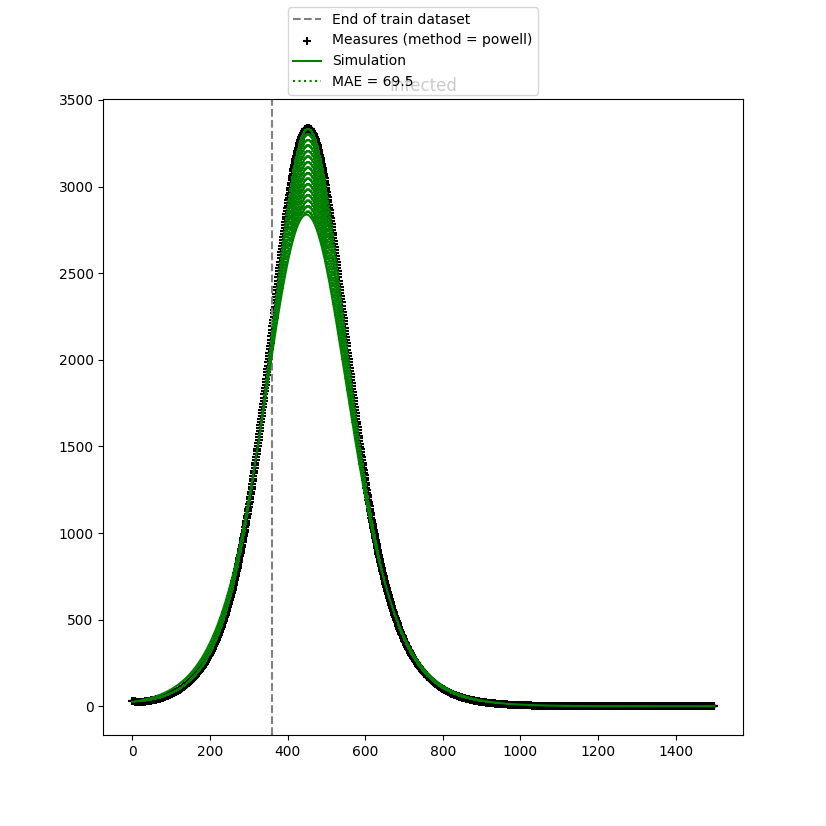}
            \caption{ Children }
        \end{subfigure} 
        \begin{subfigure}[b]{0.3\textwidth}
            \includegraphics[width=\linewidth, bb= 0 0 826 826]{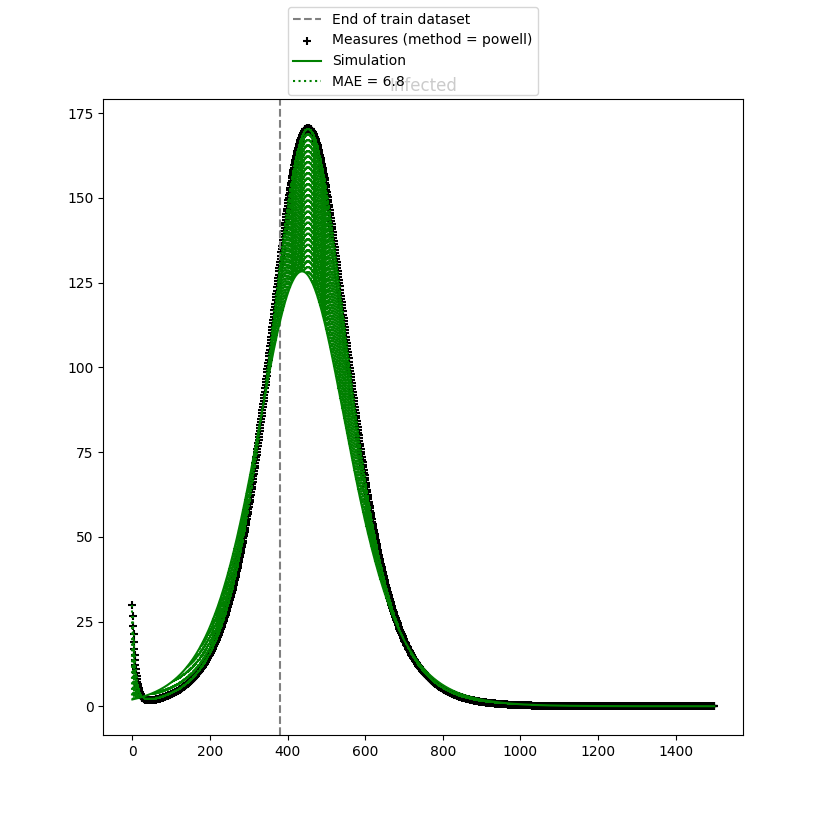}
            \caption{ Seniors }
        \end{subfigure} 
        \begin{subfigure}[b]{0.3\textwidth}
            \includegraphics[width=\linewidth, bb= 0 0 826 826]{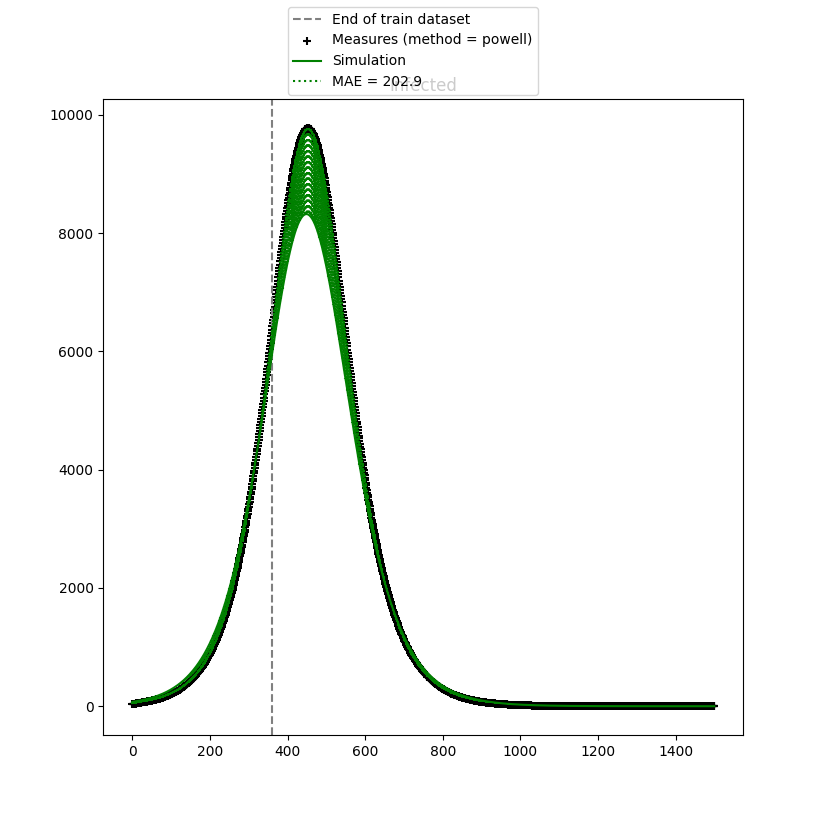}
            \caption{ Adults }
        \end{subfigure} 
    \end{subfigure} 
\caption{Best performing calibration methods for each subgroup in an \textit{age-based} population subgroup SIR model in the low data regime.}\label{fig:top-lowdata-nonoise-withsubgroups}
\end{figure}

\begin{figure}[ht]
\centering
    \begin{subfigure}[b]{\textwidth}
        \centering
        \begin{subfigure}[b]{0.3\textwidth}
            \includegraphics[width=\linewidth, bb= 0 0 826 826]{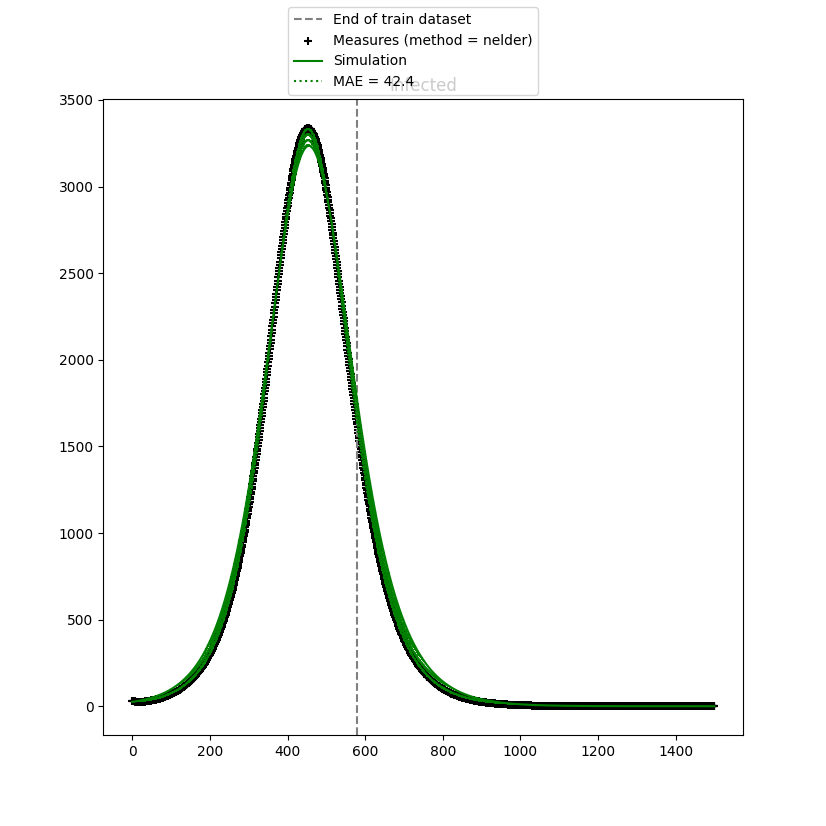}
            \caption{ Children }
        \end{subfigure} 
        \begin{subfigure}[b]{0.3\textwidth}
            \includegraphics[width=\linewidth, bb= 0 0 826 826]{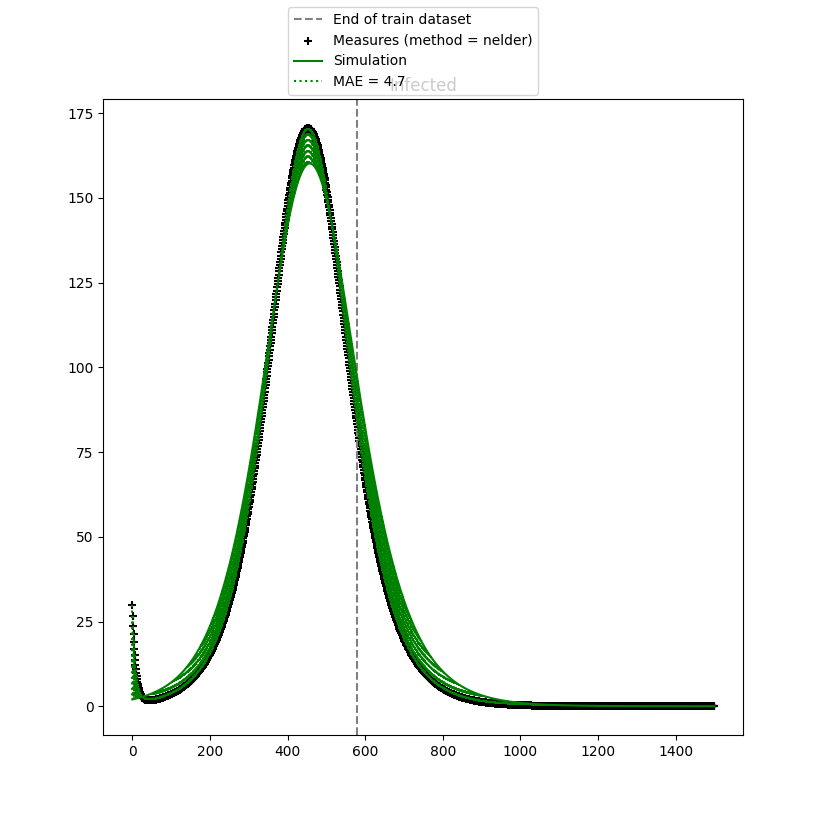}
            \caption{ Seniors }
        \end{subfigure} 
        \begin{subfigure}[b]{0.3\textwidth}
            \includegraphics[width=\linewidth, bb= 0 0 826 826]{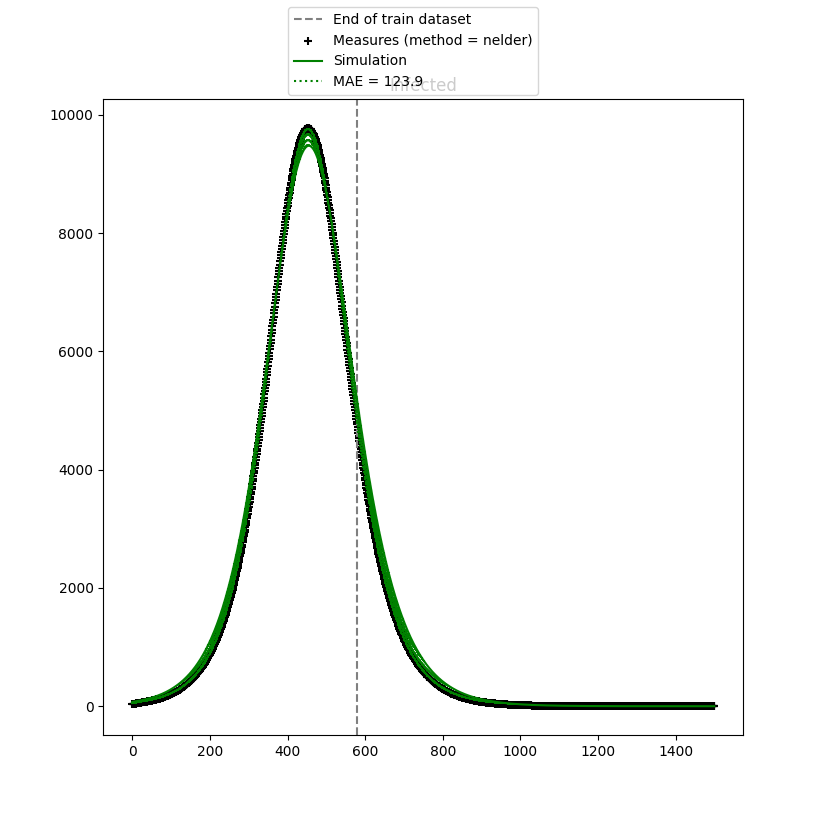}
            \caption{ Adults }
        \end{subfigure} 
    \end{subfigure} 
\caption{Best performing calibration methods for each subgroup in an \textit{age-based} population subgroup SIR model in the high data regime.}\label{fig:top-highdata-nonoise-withsubgroups}
\end{figure}

As can be seen in Figure ~\ref{fig:top-lowdata-nonoise-withsubgroups} Powell's method performs the best for all subgroups in the low data regime. On the other hand, when given more data, the Nelder-mead optimization method performs the best (Figure ~\ref{fig:top-highdata-nonoise-withsubgroups}). It is interesting to note that these methods also seemed quite successful in the simpler experiments discussed in the previous chapters. 

\begin{figure}[ht]
\centering
    \begin{subfigure}[b]{\textwidth}
        \centering
        \begin{subfigure}[b]{0.3\textwidth}
            \includegraphics[width=\linewidth, bb= 0 0 826 826]{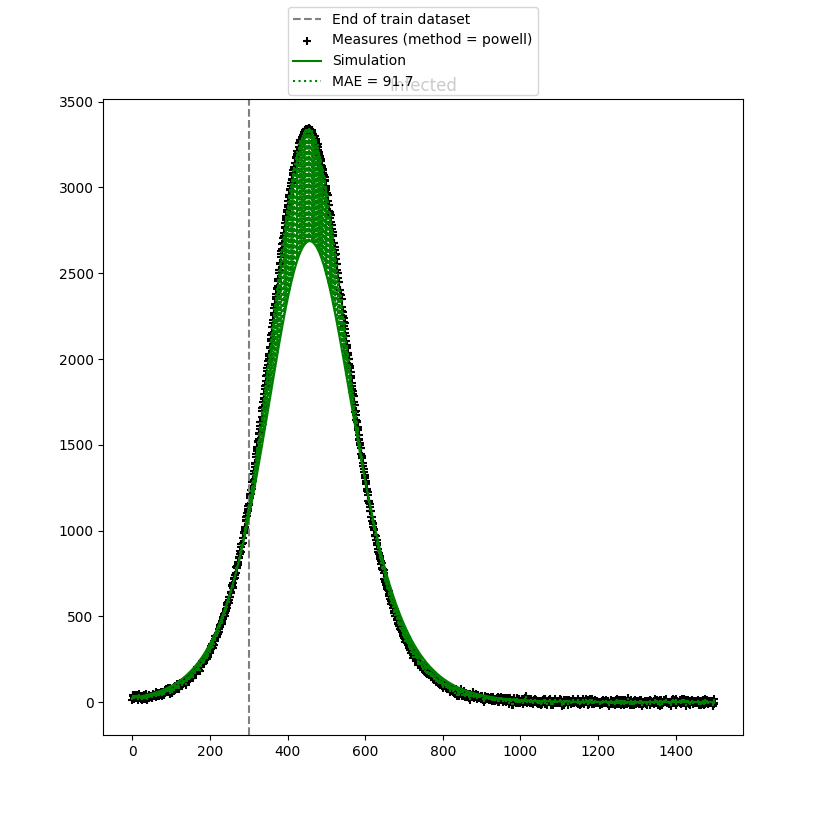}
            \caption{ Children }
        \end{subfigure} 
        \begin{subfigure}[b]{0.3\textwidth}
            \includegraphics[width=\linewidth, bb= 0 0 826 826]{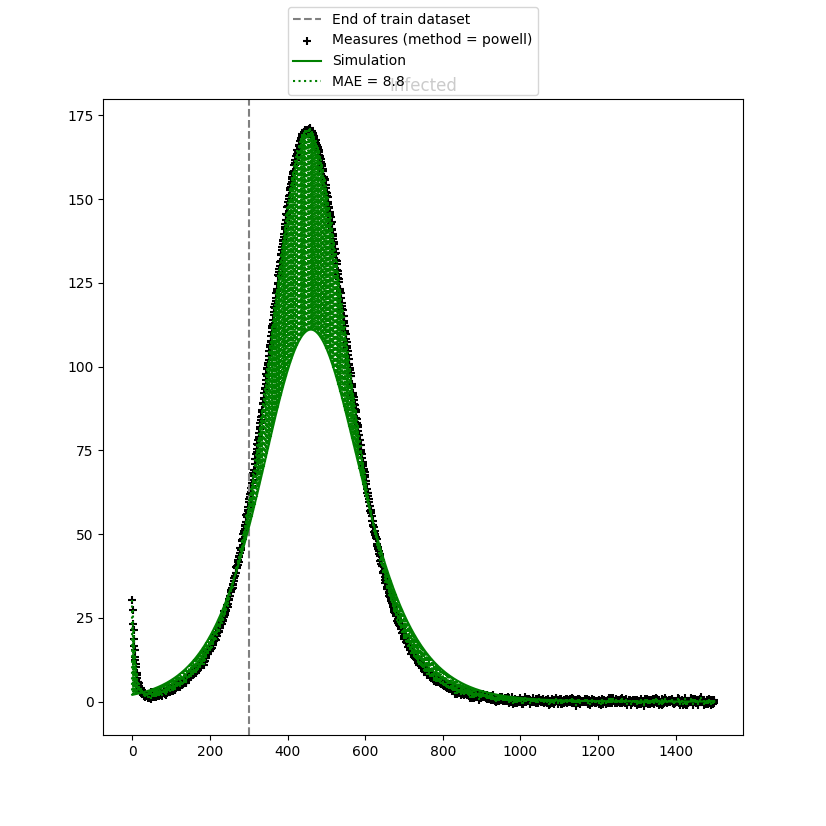}
            \caption{ Seniors }
        \end{subfigure} 
        \begin{subfigure}[b]{0.3\textwidth}
            \includegraphics[width=\linewidth, bb= 0 0 826 826]{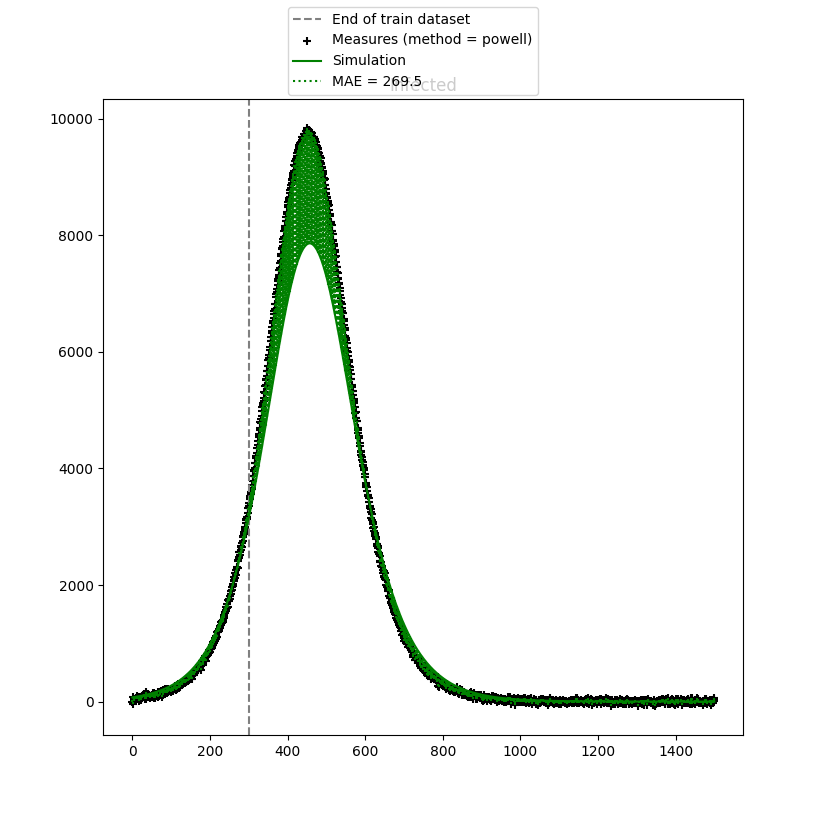}
            \caption{ Adults }
        \end{subfigure} 
    \end{subfigure} 
\caption{Best performing calibration methods for each subgroup in an \textit{age-based} population subgroup SIR model with added \textbf{noise} in the low data regime.}\label{fig:top-lowdata-withnoise-withsubgroups}
\end{figure}

\begin{figure}[ht]
\centering
    \begin{subfigure}[b]{\textwidth}
        \centering
        \begin{subfigure}[b]{0.3\textwidth}
            \includegraphics[width=\linewidth, bb= 0 0 826 826]{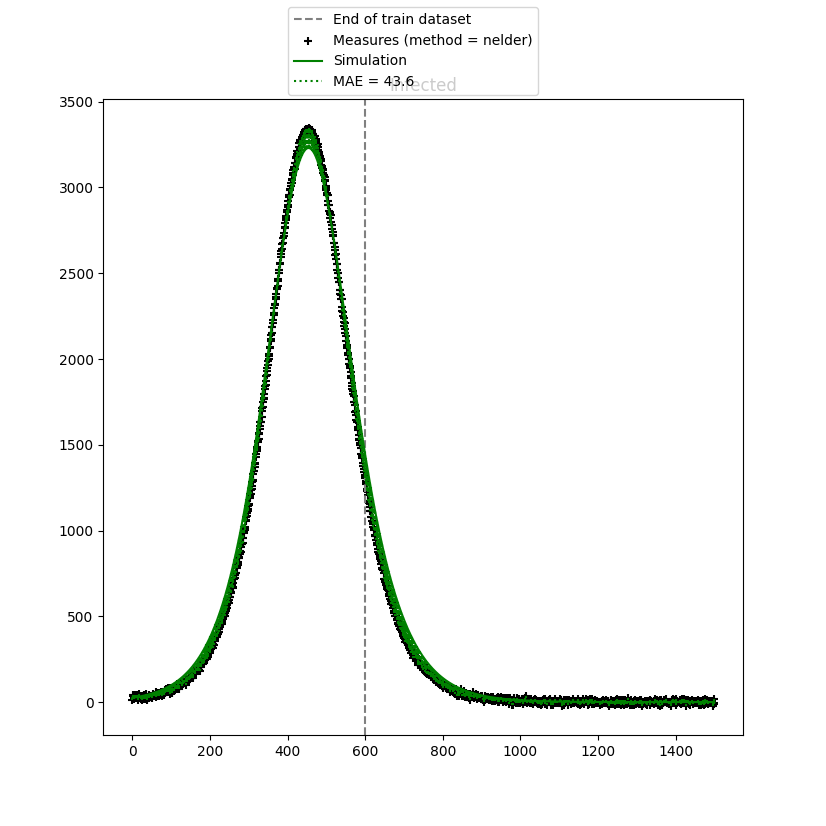}
            \caption{ Children }
        \end{subfigure} 
        \begin{subfigure}[b]{0.3\textwidth}
            \includegraphics[width=\linewidth, bb= 0 0 826 826]{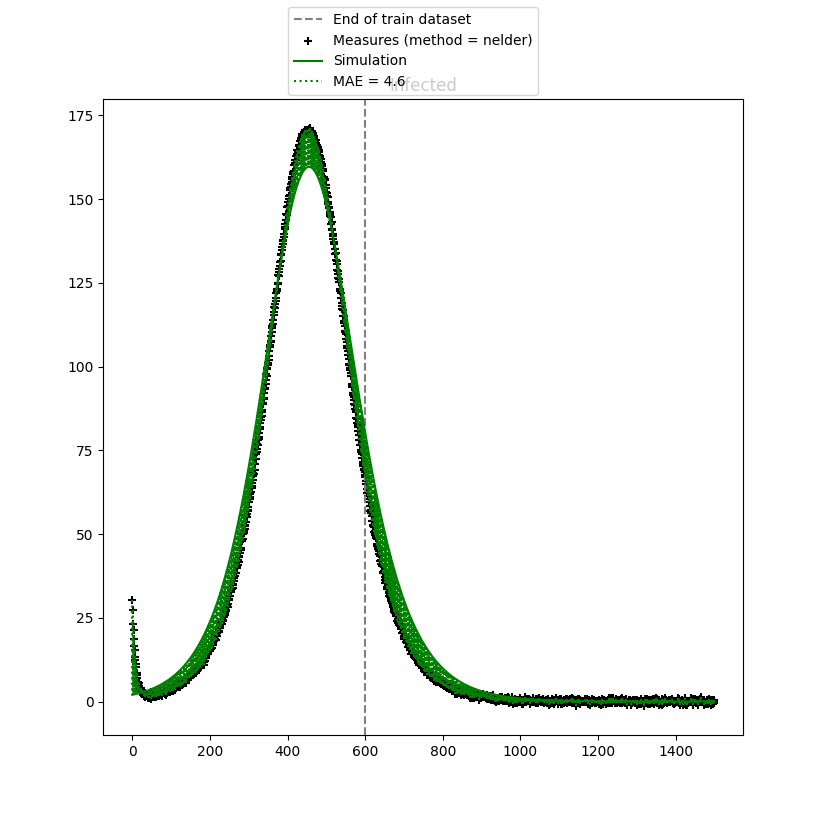}
            \caption{ Seniors }
        \end{subfigure} 
        \begin{subfigure}[b]{0.3\textwidth}
            \includegraphics[width=\linewidth, bb= 0 0 826 826]{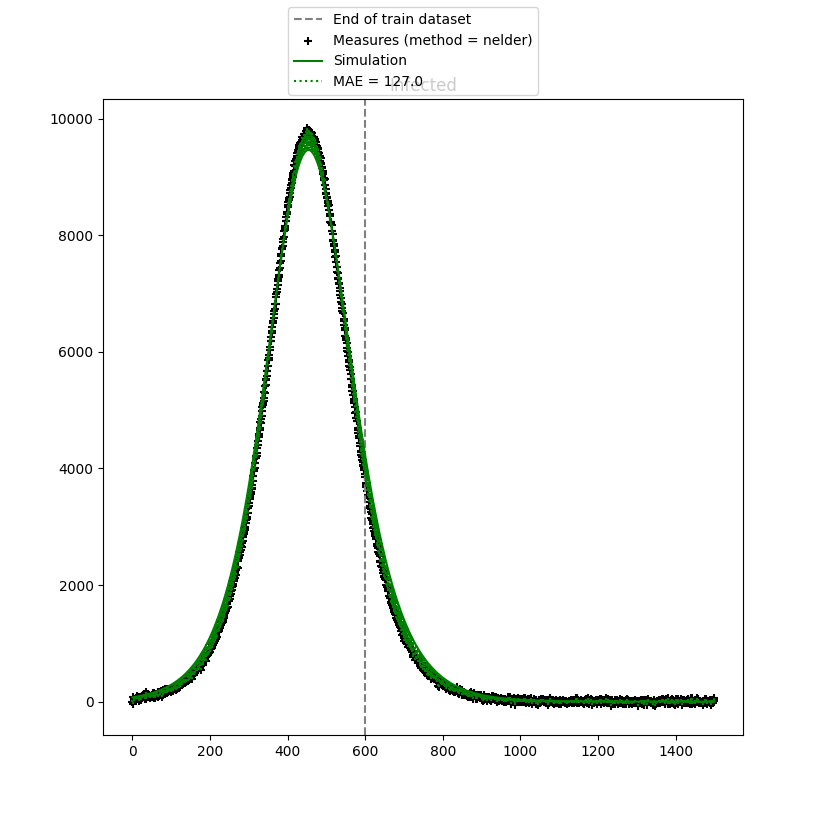}
            \caption{ Adults }
        \end{subfigure} 
    \end{subfigure} 
\caption{Best performing calibration methods for each subgroup in an \textit{age-based} population subgroup SIR model with added \textbf{noise} in the high data regime.}\label{fig:top-highdata-withnoise-withsubgroups}
\end{figure}

Next, we also investigate whether these methods keep up their performance when we add noise to the data. In the low-data regime, Powell's method still performs the best for all population subgroups. The next best was the truncated Newton's method but it worked well only for subgroup 2 and was not up to the mark in other subgroups. We promote using Powell's method in the low data regime even for noisy data when age-based population subgroups are considered (Figure ~\ref{fig:top-lowdata-withnoise-withsubgroups}). On the other hand, in the high data regime, we witnessed more than one winner for each subgroup. Basinhopping optimization performed the best for the first subgroup, immediately followed by Nelder-Mead's algorithm (similar MAE). Powell's method wins in the second group, however, again followed by Nelder-Mead's algorithm with a very similar MAE. And, for the third subgroup, Nelder-Mead performs the best and is followed by Basinhopping and Powell (with a comparatively much higher MAE). All in all, it would be best to suggest using Nelder-Mead's algorithm in the high data regime for noisy age-based population subgroups' data (Figure ~\ref{fig:top-highdata-withnoise-withsubgroups}). We again note here that even though the fit shown in Figure ~\ref{fig:top-highdata-withnoise-withsubgroups} looks good, it is due to a large amount of data being available to the calibration methods for training. This is not very practical in real-world scenarios but is good for a research investigation. 

In the next chapter, we make an attempt to draw some practical conclusions about which calibration algorithm to use in different cases from all of these discussed experiments.

%% file: chapters/05_practical_conclusions.tex
Through the various experiments and results discussed in the previous chapters, we now discuss some lessons and practical conclusions that can be drawn from the use of optimization methods for epidemic model parameter calibration. 

\paragraph{No single optimization method is universally superior.} These strategies appear to be highly dependent on particular characteristics of the epidemic model and data that is accessible. Different methods have different strengths and weaknesses, and the best choice needs to be determined by the problem at hand. 

\paragraph{Calibration or parameter estimation is a difficult problem.} Epidemic models often involve a large number of parameters, making parameter estimation difficult. Moreover, even with a small number of parameters, the performance of various methods varies highly depending on the amount of data available for training, noise levels, model complexities, and consideration of commonly unaccounted factors such as contact matrices for population subgroups. Any outcomes and uncertainties associated with the estimates need to be thoroughly validated. 

\paragraph{The quality of the data is crucial.} The accuracy of calibration depends on the quality of the data deployed for estimation. Poor quality data, such as incomplete or inconsistent data, or noisy data, can lead to erroneous estimations and hinder the model's ability to make accurate predictions. 

\begin{figure}
    \centering
    \includegraphics[width=\linewidth]{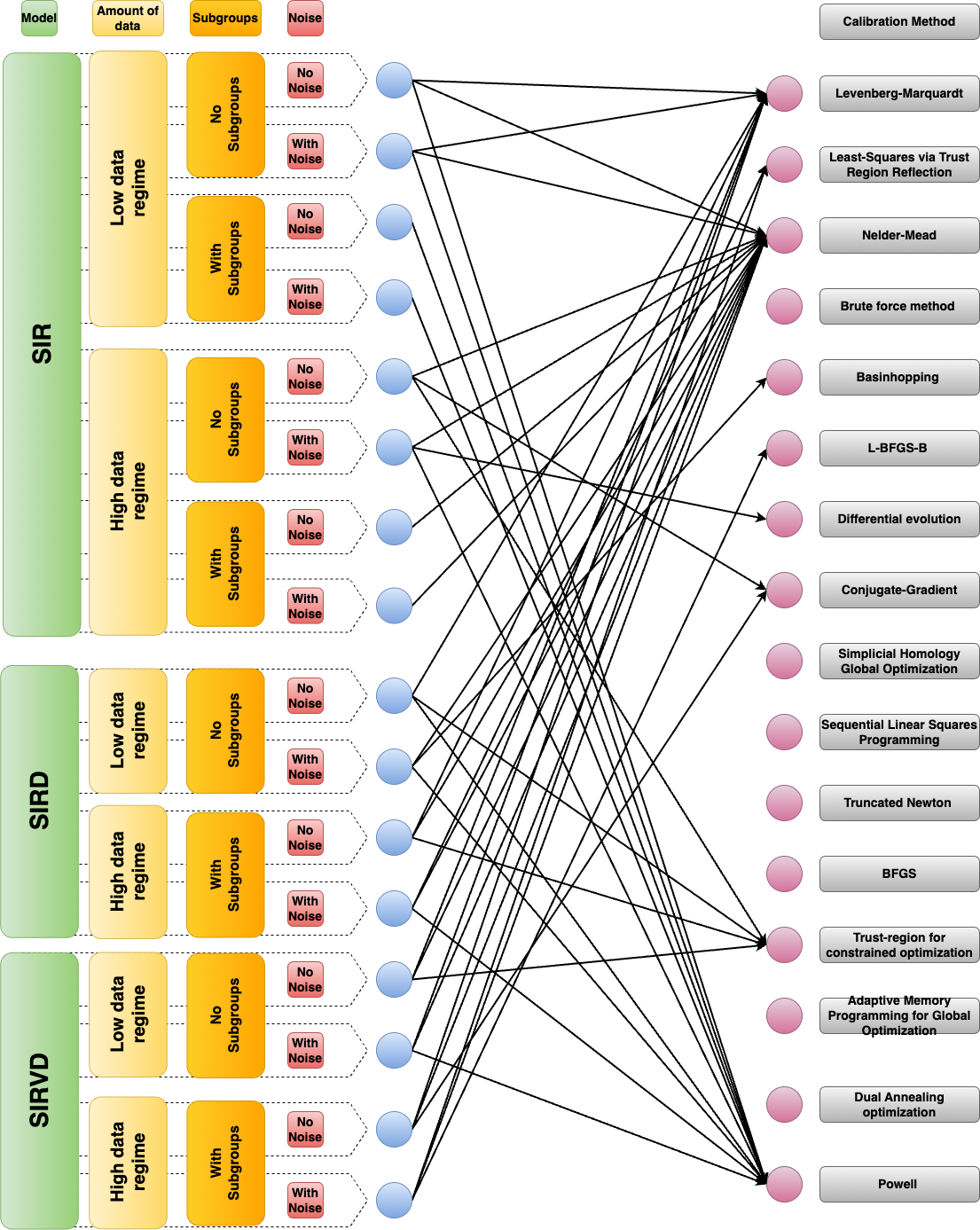}
    \caption{Practical Conclusions: A bipartite graph to observe practical conclusions for choosing an appropriate optimization algorithm to calibrate a compartmental epidemiological model (SIR/SIRD/SIRVD) considering the amount of data available for training, noise in data, and population subgroups. A tabular version of the bipartite graph can be found in the appendix (Section \ref{sec:tabular_bipartite}). 
    } 
    \label{fig:bipartite_graph}
\end{figure}

To summarize, optimization methods can be a strong tool for epidemic model calibration, but their application requires careful evaluation of the epidemic model's specific properties as well as the available data. Nevertheless, in Figure ~\ref{fig:bipartite_graph} we visualize a bipartite graph to connect the best choice of calibration methods with (1) model complexity (SIR, SIRD, SIRVD), (2) amount of data (low data regime, high data regime), (3) presence of noise, and (4) consideration of population subgroups. We note here that, the methods train on the first few days from the data (based on the training data regime --- low or high), but we evaluate them by their prediction on the entire dataset, i.e., the remaining days (not available for training) act as the testing data for evaluation. A tabular version of the bipartite graph can be found in the appendix (Section \ref{sec:tabular_bipartite}).

%% file: chapters/06_reinforcement_learning.tex
\subsection{Motivation} 

Reinforcement learning (RL) can be a useful technique for calibrating epidemiological models. RL is a branch of machine learning that deals with agents learning from trial-and-error interactions with an environment to maximize a reward signal. In the context of epidemiological modeling, the environment could be a simulation of the spread of disease, and the reward signal could be based on how closely the output of the simulation matches real-world data. The RL agent would adjust the parameters of the model during each iteration to try to optimize this reward signal. 

RL can explore a wide range of parameter values without the need to specify a prior distribution. This is particularly useful as the epidemiological dynamics are often complex and difficult to model explicitly. It also assists in identifying portions of the parameter space that are important for successful modeling. RL can also be calibrated in real-time, which means it can be constantly updated as new data becomes available. This is especially essential in our application because making informed choices require accurate and up-to-date models. 

However, it's important to note that using RL for model calibration requires careful consideration of the reward signal and the exploration strategy. The calibrated model needs to be assessed using independent data to ensure that it appropriately depicts the disease's underlying dynamics. Furthermore, it demands large amounts of data, which in some cases can be difficult to obtain, and there is a risk of overfitting, in which the model learns to fit the noise in the data rather than the underlying trends. 

\subsection{Idea} 


\paragraph{Problem Definition.} The specific goal of RL is to find the best set of model parameters starting from an initial guess of their values. In other words, given an epidemiological compartmental model, with specified parameters to be calibrated, RL's target is to start from an initial guess of parameter values, and sequentially update them to locate better values in terms of MAE between the predicted and true data points of the \textit{Infected} compartment. 

\paragraph{RL agent design.} We consider designing our RL agent using the Proximal Policy Optimization (PPO) algorithm for epidemiological compartmental model calibration. PPO is a popular RL algorithm that learns a parameterized policy function that maps states to actions. 

We define the policy of the agent, $\pi$(a|s, $\theta$), as a probability distribution over actions \textit{a} given state \textit{s} and parameters $\theta$. PPO updates the policy using the following rule: 

\[ L(\theta) = \mathbb{E}\Big[\text{min}\Big(\frac{\pi_\theta(a|s)}{p_\theta(a|s)}A(s,a), clip\Big(\frac{\pi_\theta(a|s)}{p_\theta(a|s)}, 1-\epsilon, 1+\epsilon\Big)A(s,a)\Big)\Big] \]

where $\theta$ represents the parameters of the policy neural network, $\pi_\theta(a|s)$ is the probability of taking action $a$ in state $s$ according to the policy, $p_\theta(a|s)$ is the probability of taking action $a$ in state $s$ according to the old policy, $A(s,a)$ is the advantage function that estimates the expected return of taking action $a$ in state $s$, $\epsilon$ is a hyperparameter that controls the magnitude of the clipping, and $\mathbb{E}$ denotes the expectation over the state-action pairs encountered during training. 

\paragraph{States.} We generate a vector of random initial guesses in the range of [0, 1] corresponding to all the epidemiological model parameters. This would be the starting point of the RL agent which then will take actions to update this vector in the direction of decreasing MAE of the predicted \textit{Infected} curve. 

\paragraph{Actions.} We  defined the actions in the form of 

\begin{center}
    [\textit{update}] [\textit{parameter}] [\textit{by value}] 
\end{center}

For example, ``increase $\beta$ by 0.1", ``decrease $\gamma$ by 0.01" or ``no change". In this way, the RL agent would stay close to the initial guesses and would sequentially search for an improved set of parameters starting from the initial guess of parameter values. This is very useful when we start with the results of best-performing optimization methods (discussed in previous chapters) as initial guesses and want RL to further improve the MAE by pushing for improved estimations of model parameters instead of searching the whole space from scratch. Note that our defined action space is discrete. 

\paragraph{Reward Signal.} As feedback to the RL agent, we calculate the MAE between the predicted data points (using the estimated parameters by agent) and the true data points from the epidemiological compartmental model (we used only the \textit{Infected} compartment data points to drive the training and evaluation). 

The MAE loss is given by:

\[ R(s, a) = -\frac{1}{n}\sum_{i=1}^n |\hat{y}_i - y_i| \] 

where $n$ is the number of data points, $\hat{y}_i$ is the predicted value of the $i$-th data point, $y_i$ is the actual value of the $i$-th data point, and $s$, $a$, and $s'$ are the current state, action, and next state, respectively. The negative sign is used to convert the loss into a reward so that higher rewards correspond to lower losses. We used the MAE as a metric for model accuracy as there could be many outliers in epidemiological data and the absolute difference between the predicted and actual values is more important than the squared difference. 

\begin{figure}[ht]
    \centering
    \includegraphics[width=\linewidth, bb= 0 0 721 595]{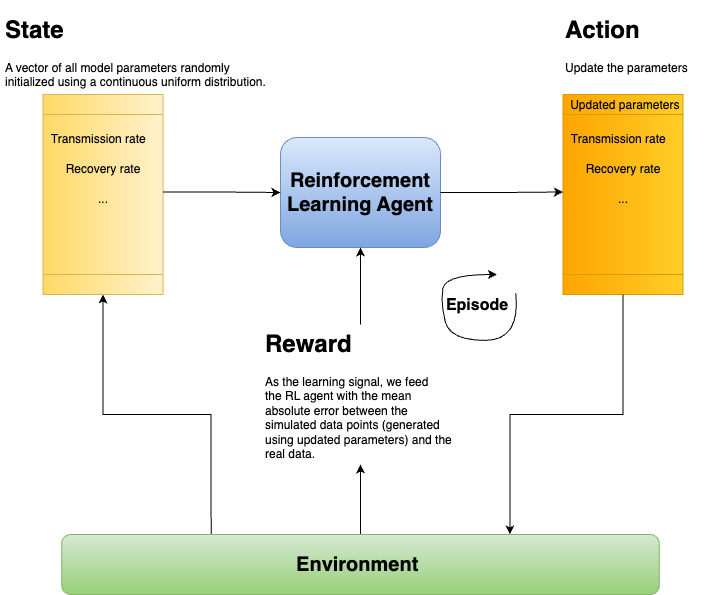}
    \caption{Reinforcement Learning based calibration of epidemiological compartmental models.} 
    \label{fig:rl}
\end{figure}

\paragraph{Episodes.} Now that we have defined the states, action space, and reward signal, we can describe the step-by-step functioning in an episode. For a given epidemiological model, at the beginning of an episode, the environment generates a random vector containing values for all the model parameters. The RL agent uses this vector as a state and takes action on this, consequently updating it to a new set of values. This updated vector now goes to the environment which then simulates an \textit{Infected} curve using these new parameter values. It then calculates the MAE between these predicted data points and the true data from the epidemiological model data. This MAE is used to construct the reward function as described above which is then sent as feedback to the RL agent by the environment to learn and update its policy. This cycle continues until a threshold of prespecified low MAE value or a prespecified number of timesteps has been reached. Figure~\ref{fig:rl} shows the idea and workflow of our RL approach. 

\subsection{Discussion} 

While deep reinforcement learning (DRL) has several advantages for calibrating epidemiological models, there are also some potential problems that need to be considered: 

\paragraph{Data requirements.} As mentioned earlier, RL requires large amounts of data to train effectively. It is undeniably difficult to get sufficient data in epidemiological modeling due to a number of problems such as limited testing capacity, a lack of dependable reporting channels, and regional variability in data quality. Furthermore, data collected during an outbreak may be biased and may not accurately reflect the disease's underlying dynamics.  

\paragraph{Overfitting.} RL models are prone to overfitting, where the model learns to fit the noise in the data rather than the underlying trends. This can lead to poor generalization performance and inaccurate model predictions.

\paragraph{Computational requirements.} RL models can be computationally expensive to train and evaluate. This can limit their practical use in some applications. 

\paragraph{Interpretability.} It can be difficult to interpret RL models, particularly when using deep neural networks as function approximators. This can make it challenging to understand the underlying mechanisms of the model and make informed decisions based on the model's outputs. 

\paragraph{Unforeseen outcomes.} RL models can be unpredictable and may produce unexpected outcomes, particularly in complex or uncertain environments, which is a huge concern in epidemiological modeling, where model predictions can have significant real-world consequences. 

Overall, while RL may have the potential to improve the accuracy and adaptability of epidemiological models, one must carefully consider the potential problems. Therefore, further research is needed to fully understand the benefits and limitations of using DRL for epidemic model calibration.

%% file: chapters/08_conclusion.tex
Here are a few of our final thoughts from this study on calibrating epidemiological models: 

\paragraph{The choice of model is critical.} There are numerous types of epidemic models, and the model used might have a considerable impact on the analysis's conclusions. It is important to carefully assess which sort of model is best appropriate for the specific epidemic under study, taking into account criteria such as mode of transmission, illness history, and accessible data. 

\paragraph{Collaboration is necessary.} Epidemic modeling is a complicated and interdisciplinary area, and accurate calibration of epidemic models frequently necessitates collaboration among experts in epidemiology, mathematics, statistics, and computer science. Researchers can pool their skills and produce more accurate and robust models by collaborating, for example, when initial expert estimates of parameters are good, they can be shared and confirmed by all branches. 

\paragraph{Sensitivity analysis is important.} Understanding which parameters have the greatest impact on the model output can help to prioritize future data collection efforts. 

\paragraph{There is no universal winner of calibration methods.} It is critical to carefully analyze the benefits and drawbacks of multiple methods before selecting one for a problem at hand. 

\paragraph{Any result should be communicated clearly.} The calibration analysis results should be conveyed to stakeholders, including policymakers and the general public, in a clear and transparent manner. Researchers can help to develop faith in the model and the analysis by presenting the results in a clear and intelligible manner, facilitating informed decision-making. 

\paragraph{The analysis should be updated periodically.} Epidemics are dynamic and evolve over time, hence epidemic model analyses should be often updated to reflect the most recent data and knowledge. This will enable researchers to provide the most accurate and up-to-date information to stakeholders and decision-makers.

%% file: acknowledge.tex
The authors would like to acknowledge the support of the U.S. National Science Foundation grants 1840761,  1934388, and 1840761, the U.S. National Institutes of Health grant 5R01GM121753, New York University Abu Dhabi Center for Interacting Urban Networks (CITIES), and NYU Wireless. We also acknowledge \textit{Clément Mauget} and \textit{Roxane Leduc} for their initial work on this topic as part of a summer internship with Professor Dennis Shasha and \textit{Vikramsreehari Mullachery} for a detailed review and comments. 

%% file: chapters/10_appendix.tex
\section*{Tabular form of practical conclusions (bipartite graph)} 
\label{sec:tabular_bipartite} 

Table \ref{table:bipartite_table} shows the practical conclusions (bipartite graph) in a tabular form. 

\begin{sidewaystable}[ht]
\caption{Tabular form of practical conclusions (bipartite graph)}
\label{table:bipartite_table}
\centering
\begin{tabular}{|c|c|c|c|ccc|}
\hline
\textbf{\begin{tabular}[c]{@{}c@{}}Model \\ Complexity\end{tabular}} & \textbf{\begin{tabular}[c]{@{}c@{}}Amount of Data \\ for training\end{tabular}} & \textbf{\begin{tabular}[c]{@{}c@{}}Presence of \\ Noise\end{tabular}} & \textbf{\begin{tabular}[c]{@{}c@{}}Population \\ subgroups \\ considered?\end{tabular}} & \multicolumn{3}{c|}{\textbf{Top 3 calibration methods}}                                                                                                          \\ \hline
\multirow{8}{*}{\textbf{SIR}}                                        & \multirow{4}{*}{Low}                                                            & No                                                                    & No                                                                                      & \multicolumn{1}{c|}{Nelder-Mead}   & \multicolumn{1}{c|}{Powell}        & Least-squares                                                                          \\ \cline{3-7} 
                                                                     &                                                                                 & Yes                                                                   & No                                                                                      & \multicolumn{1}{c|}{Nelder-Mead}   & \multicolumn{1}{c|}{Least-squares} & Powell                                                                                 \\ \cline{3-7} 
                                                                     &                                                                                 & No                                                                    & Yes                                                                                     & \multicolumn{3}{c|}{Powell}                                                                                                                                      \\ \cline{3-7} 
                                                                     &                                                                                 & Yes                                                                   & Yes                                                                                     & \multicolumn{3}{c|}{Powell}                                                                                                                                      \\ \cline{2-7} 
                                                                     & \multirow{4}{*}{High}                                                           & No                                                                    & No                                                                                      & \multicolumn{1}{c|}{CG}            & \multicolumn{1}{c|}{Nelder-Mead}   & \begin{tabular}[c]{@{}c@{}}Trust-region based \\ constrained optimization\end{tabular} \\ \cline{3-7} 
                                                                     &                                                                                 & Yes                                                                   & No                                                                                      & \multicolumn{1}{c|}{Nelder-Mead}   & \multicolumn{1}{c|}{Powell}        & Differential Evolution                                                                 \\ \cline{3-7} 
                                                                     &                                                                                 & No                                                                    & Yes                                                                                     & \multicolumn{3}{c|}{Nelder-Mead}                                                                                                                                 \\ \cline{3-7} 
                                                                     &                                                                                 & Yes                                                                   & Yes                                                                                     & \multicolumn{3}{c|}{Nelder-Mead}                                                                                                                                 \\ \hline
\textbf{}                                                            &                                                                                 &                                                                       &                                                                                         & \multicolumn{1}{c|}{}              & \multicolumn{1}{c|}{}              &                                                                                        \\ \hline
\multirow{4}{*}{\textbf{SIRD}}                                       & \multirow{2}{*}{Low}                                                            & No                                                                    & No                                                                                      & \multicolumn{1}{c|}{Powell}        & \multicolumn{1}{c|}{Least-squares} & \begin{tabular}[c]{@{}c@{}}Trust-region based \\ constrained optimization\end{tabular} \\ \cline{3-7} 
                                                                     &                                                                                 & Yes                                                                   & No                                                                                      & \multicolumn{1}{c|}{Basinhopping}  & \multicolumn{1}{c|}{Nelder-Mead}   & Powell                                                                                 \\ \cline{2-7} 
                                                                     & \multirow{2}{*}{High}                                                           & No                                                                    & No                                                                                      & \multicolumn{1}{c|}{Least-squares} & \multicolumn{1}{c|}{Nelder-Mead}   & \begin{tabular}[c]{@{}c@{}}Trust-region based \\ constrained optimization\end{tabular} \\ \cline{3-7} 
                                                                     &                                                                                 & Yes                                                                   & No                                                                                      & \multicolumn{1}{c|}{Least-squares} & \multicolumn{1}{c|}{Nelder-Mead}   & Powell                                                                                 \\ \hline
\textbf{}                                                            &                                                                                 &                                                                       &                                                                                         & \multicolumn{1}{c|}{}              & \multicolumn{1}{c|}{}              &                                                                                        \\ \hline
\multirow{4}{*}{\textbf{SIRVD}}                                      & \multirow{2}{*}{Low}                                                            & No                                                                    & No                                                                                      & \multicolumn{1}{c|}{Nelder-Mead}   & \multicolumn{1}{c|}{Least-squares} & \begin{tabular}[c]{@{}c@{}}Trust-region based \\ constrained optimization\end{tabular} \\ \cline{3-7} 
                                                                     &                                                                                 & Yes                                                                   & No                                                                                      & \multicolumn{1}{c|}{Least-squares} & \multicolumn{1}{c|}{Nelder-Mead}   & Powell                                                                                 \\ \cline{2-7} 
                                                                     & \multirow{2}{*}{High}                                                           & No                                                                    & No                                                                                      & \multicolumn{1}{c|}{CG}            & \multicolumn{1}{c|}{Least-squares} & Nelder-Mead                                                                            \\ \cline{3-7} 
                                                                     &                                                                                 & Yes                                                                   & No                                                                                      & \multicolumn{1}{c|}{Nelder-Mead}   & \multicolumn{1}{c|}{Least-squares} & L-BFGS-B                                                                               \\ \hline
\end{tabular}
\end{sidewaystable}